\pdfoutput=1

\documentclass[11pt]{article}

\usepackage[table]{xcolor}
\usepackage[final]{acl}

\usepackage{times}
\usepackage{latexsym}

\usepackage[T1]{fontenc}

\usepackage[utf8]{inputenc}

\usepackage{microtype}

\usepackage{inconsolata}

\usepackage{graphicx}

\usepackage{comment}
\usepackage{todonotes}
\usepackage{booktabs}
\usepackage{adjustbox}
\usepackage{multirow}
\usepackage{array}
\usepackage{rotating}
\usepackage{amssymb}
\usepackage{gb4e}
\noautomath

\title{TurBLiMP: A Turkish Benchmark of Linguistic Minimal Pairs}

\author{
Ezgi Ba\c{s}ar \quad
Francesca Padovani \quad
Jaap Jumelet \quad
Arianna Bisazza \\
Center for Language and Cognition (CLCG), University of Groningen \\
\texttt{\{e.basar, f.padovani, j.w.d.jumelet, a.bisazza\}@rug.nl}
}

\begin{document}
\maketitle
\begin{abstract}
We introduce TurBLiMP, the first Turkish benchmark of linguistic minimal pairs, designed to evaluate the linguistic abilities of monolingual and multilingual language models (LMs). Covering 16 linguistic phenomena with 1000 minimal pairs each, TurBLiMP fills an important gap in linguistic evaluation resources for Turkish. In designing the benchmark, we give extra attention to two properties of Turkish that remain understudied in current syntactic evaluations of LMs, namely word order flexibility and subordination through morphological processes. Our experiments on a wide range of LMs and a newly collected set of human acceptability judgments reveal that even cutting-edge Large LMs still struggle with grammatical phenomena that are not challenging for humans, and may also exhibit different sensitivities to word order and morphological complexity compared to humans.

\end{abstract}

\section{Introduction}

A foundational insight in linguistics research is that applying minimal changes to a sentence can render it entirely acceptable or unacceptable to native speakers \citep{Chomsky1965-CHOAOT-2}. Minimal pairs, as illustrated in Example (\ref{ex:1}), are a widely used diagnostic tool in linguistics.

\begin{exe}
\ex \begin{xlist}
        \ex[]{People in Istanbul love cats.}
        \ex[*]{People in Istanbul loves cats.}
    \end{xlist}
\label{ex:1}
\end{exe}

Minimal pairs have been a cornerstone of linguistic analysis for decades, and in recent years they have become a vital tool for the linguistic evaluation of language models (LMs).
\citet{warstadt-etal-2020-blimp-benchmark} published the first large-scale English \textbf{B}enchmark of \textbf{Li}nguistic \textbf{M}inimal \textbf{P}airs (BLiMP) in an effort to systematically evaluate the linguistic knowledge of language models, and since then various benchmarks have been introduced for other languages. 

We contribute to this growing collection by introducing the first Turkish benchmark of linguistic minimal pairs. 
TurBLiMP enriches the typological diversity of available linguistic evaluation benchmarks by incorporating a morphologically rich agglutinative language with highly flexible word order.
While Turkish and other agglutinative languages like Finnish have been the object of several studies focusing on word-level morphology \citep{ismayilzada-etal-2025-evaluating}, the effects of word order flexibility and morphological complexity on the robustness of sentence-level grammatical judgments have not been studied in detail before. We fill this gap by introducing two sets of experimental minimal pair paradigms. 

Our evaluation shows that even top-performing LMs suffer performance losses under word order or subordination manipulations, revealing sensitivities that would otherwise go undetected. Compared to the acceptability judgments we collected from native speakers, baseline tests across 13 models and 16 Turkish phenomena demonstrate that Large LMs can struggle with linguistic tasks where humans perform reliably. By providing this resource, we aim to facilitate linguistically motivated NLP research and contribute a high-quality dataset for linguists and NLP researchers.

\section{Minimal Pair Benchmarks}
Minimal pairs have played an important role for evaluating the linguistic abilities of language models, targeting phenomena such as subject-verb agreement \citep{linzen-etal-2016-assessing}, filler-gap dependencies \citep{wilcox-etal-2018-rnn}, and negative polarity items \citep{jumelet-hupkes-2018-language}.
\citet{warstadt-etal-2020-blimp-benchmark} then established an English benchmark of 67,000 sentence pairs testing 67 paradigms through automated generation based on linguist-curated templates. 
This work inspired numerous adaptations for other languages, each employing different benchmark creation strategies. 
Benchmarks using a similar template-based approach as BLiMP include CLiMP \citep[Chinese,][]{xiang-etal-2021-climp}, ZhoBLiMP \citep[Chinese,][]{liu2024zhoblimpsystematicassessmentlanguage}, BLiMP-NL \citep[Dutch,][]{10.1162/coli_a_00559}, and for Basque/Swahili/Hindi by \citet{kryvosheieva-levy-2025-controlled}.
Another approach is based on modifying Universal Dependency trees, which has been used by SLING \citep[Chinese,][]{song-etal-2022-sling}, RuBLiMP \citep[Russian,][]{taktasheva-etal-2024-rublimp}, and MultiBLiMP \citep{jumelet2025multiblimp10massivelymultilingual}, a multilingual benchmark covering 101 languages.
Other approaches include the extraction of minimal pairs from linguistics journals, employed by JBLiMP \citep[Japanese,][]{someya-oseki-2023-jblimp}, manual creation of pairs, as done for Icelandic by \citet{armannsson-etal-2025-icelandic}, and the usage of LLMs for generating pairs, as done for Tamil and Indonesian by \citet{leong2023bhasaholisticsoutheastasian}.

Methodological innovations across these benchmarks reveal key trade-offs between scale, linguistic coverage, and data quality. Template-based generation enables large datasets but risks producing unnatural sentences \citep{javier-vazquez-martinez-etal-2023-evaluating}, while manual extraction from literature or learner corpora ensures quality at the cost of scale. Some of the benchmarks incorporate hybrid approaches and human validation steps to balance these concerns. TurBLiMP too is the result of such hybrid approaches. While creating our benchmark, we developed strategies specifically adapted to the challenges of creating minimal pairs for Turkish.

\section{Turkish Morphosyntax \& NLP}
\label{sect:turkish-lm}
Turkish presents a particularly interesting case for BLiMP-style evaluation due to its flexible word order and rich morphological system. Turkish syntactically licenses \textbf{\textit{all} six possible orderings} of the main sentence constituents: Subject-Object-Verb (SOV) represents the canonical order, while other permutations introduce subtle pragmatic variations without altering the core meaning of the sentence.
As a result, evaluating LMs on a language like Turkish makes it possible to test them for their robustness to different positional patterns or grammatical hierarchies, in a way that is not possible with English and other fixed-order languages that dominate the training material of current LLMs.

Furthermore, Turkish has highly productive agglutinative morphology, whereby words typically consist of several morphemes attached to a root. Speakers can easily produce and understand numerous legitimate but low-frequency word forms through regular morphological processes, yielding substantially larger vocabulary requirements for LMs compared to analytic and fusional languages.
Many syntactic phenomena are realized in Turkish through morphology, rather than by separate function words like in English and other Indo-European languages that form a large chunk of the world's highest-resource languages.
A salient example is \textbf{subordination}, which largely involves the use of suffixes to nominalize or adverbialize the verb of the embedded clause.

\begin{exe}
    \ex
    \gll Elif'in Gaye’yi sev-diğ-in-i bil-iyor-um. \\
    Elif\textsc{\scriptsize -3SG.GEN} Gaye\textsc{\scriptsize -ACC} like\textsc{\scriptsize -NMLZ-3SG.POSS-ACC} know\textsc{\scriptsize -PROG-1SG} \\
    \glt `I know that Elif likes Gaye.'
\label{ex:2}

\end{exe}

For instance, the subordination structure in Example \ref{ex:2} can be intuitively conveyed as `I know the liking of Gaye by Elif'. Here, the nominalized verb `like' takes an accusative case suffix as the object of `know', but also a possessive agreement suffix corresponding to the genitive suffix taken by the subordinate subject `Elif'.

In general, agglutinative languages such as Finnish, Tamil, Basque, Indonesian or Japanese have been shown to be particularly challenging for neural models \cite{gerz-etal-2018-relation,cotterell-etal-2016-sigmorphon,park-etal-2021-morphology,arnett-bergen-2025-language}. Focusing on Turkish, \citet{Ataman2017LinguisticallyMV} established that fixed vocabulary constraints combined with suboptimal sub-word segmentation significantly impair neural machine translation performance for agglutinative languages. \citet{ismayilzada-etal-2025-evaluating} studied LLMs' ability to produce and systematically understand novel well-formed combinations of morphemes in Turkish and Finnish, and reported limited morphological generalization.
These findings suggest that studying flexible-order, morphologically rich languages like Turkish can provide unique insights into the true linguistic capabilities of LMs beyond surface fluency.

\section{TurBLiMP}

The creation of the TurBLiMP benchmark was motivated by the need for a controlled evaluation benchmark that accounts for the unique linguistic properties of Turkish. Some of these properties include flexible word order, morphological richness, optional pro-drop, and syncretism in third-person subject-verb agreement markers. 
We now provide a brief linguistic background on our minimal pairs.

\begin{table*}[ht]
\tiny
\centering
\renewcommand{\arraystretch}{1.5}
\begin{tabular}{@{}p{0.15\textwidth}@{} @{}p{0.58\textwidth}@{} @{}>{\raggedright\arraybackslash}p{0.27\textwidth}@{}}
{\small \textbf{Phenomenon}}           & {\small \textbf{Minimal pair}} &  {\small \textbf{Translation}}\\ \hline
{\scriptsize Anaphor Agreement} & 
\renewcommand{\arraystretch}{1.0}\begin{tabular}[t]{@{}l@{\hspace{2pt}}l@{\hspace{2pt}}l@{\hspace{2pt}}l@{\hspace{2pt}}l@{\hspace{2pt}}l@{\hspace{2pt}}l@{}}
\textit{Gezi} & \textit{rota-sın-ı} & \textit{[\underline{kendi-miz}} & \textit{/*\underline{kendi-niz}]} & \textit{internet-e} & \textit{bak-ma-dan} & \textit{oluştur-du-k.} \\
trip & route{\tiny \textsc{-3sg.poss-acc}} & [self{\tiny \textsc{-1pl.poss}} & /*self{\tiny \textsc{-2pl.poss}}] & internet{\tiny \textsc{-dat}} & look{\tiny \textsc{-neg-abl}} & create{\tiny \textsc{-pst-1pl}} \\
\end{tabular}
& \textit{We created the trip itinerary [ourselves / *yourselves] without checking the internet.}
\\ 
{\scriptsize Arg. Struct. Trans.}  & \renewcommand{\arraystretch}{1.0}\begin{tabular}[t]{@{}l@{\hspace{2pt}}l@{\hspace{2pt}}l@{\hspace{2pt}}l@{\hspace{2pt}}l@{}}
\textit{Eş-im-in} & \textit{[\underline{zevk-in-e}} & \textit{/*\underline{zevk-in-i}]} & \textit{çok} & \textit{güven-ir-im.} \\
spouse{\tiny-\textsc{1sg.poss-3sg.gen}} & [taste{\tiny-\textsc{3sg.poss-dat}} & /*taste{\tiny-\textsc{3sg.poss-acc}}{]} & very & trust{\tiny\textsc{-AOR-1sg}} \\ 
\end{tabular}
& \textit{I trust my wife's taste a lot.}\\
{\scriptsize Arg. Struct. Ditrans.}  & \renewcommand{\arraystretch}{1.0}\begin{tabular}[t]{@{}l@{\hspace{2pt}}l@{\hspace{2pt}}l@{\hspace{2pt}}l@{\hspace{2pt}}l@{\hspace{2pt}}l@{}}
\textit{Öğretmen} & \textit{[\underline{öğrenci-ler-e}} & \textit{/*\underline{öğrenci-ler-i}]} & \textit{yeni} & \textit{konu-yu} & \textit{anlat-tı.} \\
teacher & [student{\tiny \textsc{-pl-dat}} & /*student{\tiny \textsc{-pl-acc}}{]} & new & subject{\tiny \textsc{-acc}} & explain{\tiny \textsc{-pst}} \\ 
\end{tabular}
& \textit{The teacher explained the new topic to the students.}
\\
{\scriptsize Binding}              & 
\renewcommand{\arraystretch}{1.0}\begin{tabular}[t]{@{}l@{\hspace{2pt}}l@{\hspace{2pt}}l@{\hspace{2pt}}l@{\hspace{2pt}}l@{\hspace{2pt}}l@{}}
\textit{Yaz} & \textit{tatil-in-de} & \textit{[\underline{kendi-m-i}} & \textit{/*\underline{ben-i}]} & \textit{rahatlamış} & \textit{hissed-iyor-um.} \\
summer & holiday{\tiny \textsc{-3.poss-loc}} & [self{\tiny \textsc{-1sg-acc}} & /*me] & relaxed & feel{\tiny \textsc{-prog-1sg}} \\ 
\end{tabular}
& \textit{I feel relaxed during the summer holidays.}
\\
{\scriptsize Determiners}          & \renewcommand{\arraystretch}{1.0}\begin{tabular}[t]{@{}l@{\hspace{2pt}}l@{\hspace{2pt}}l@{\hspace{2pt}}l@{\hspace{2pt}}l@{\hspace{2pt}}l@{\hspace{2pt}}l@{\hspace{2pt}}l@{\hspace{2pt}}l@{}}
\textit{Geçen} & \textit{hafta} & \textit{tad-ı} & \textit{damağ-ım-da} & \textit{kal-an} & \textit{[\underline{bir}} & \textit{/*$\varnothing$]} & \textit{tatlı} & \textit{ye-di-m.}\\
last & week & taste{\tiny \textsc{-acc}} & palate{\tiny \textsc{-1sg.poss-loc}} & stay{\tiny \textsc{-part}} & [a & /*$\varnothing$] & dessert & eat{\tiny \textsc{-pst-1sg}} \\ 
\end{tabular}
& \textit{Last week, I ate a dessert with a taste that lingered on my tongue.}
\\
{\scriptsize Ellipsis}             & \renewcommand{\arraystretch}{1.0}\begin{tabular}[t]{@{}l@{\hspace{2pt}}l@{\hspace{2pt}}l@{\hspace{2pt}}l@{\hspace{2pt}}l@{\hspace{2pt}}l@{\hspace{2pt}}l@{\hspace{2pt}}l@{\hspace{2pt}}l@{}}
\textit{Mağaza-da} & \textit{ceket-i} & \textit{Pelin} & \textit{ve} & \textit{[\underline{pantolonu}} & \textit{\underline{Cem}} & \textit{/*\underline{Cem}} & \textit{\underline{pantolonu}]} & \textit{seç-ti.} \\
store{\tiny \textsc{-loc}} & jacket{\tiny \textsc{-acc}} & Pelin & and & trouser{\tiny \textsc{-acc}} & Cem &  &  & choose{\tiny \textsc{-pst}}\\ 
\end{tabular}
& \textit{In the store, Pelin chose the jacket and Cem chose the pants.}
\\
{\scriptsize Irregular Forms}      & \renewcommand{\arraystretch}{1.0}\begin{tabular}[t]{@{}l@{\hspace{2pt}}l@{\hspace{2pt}}l@{\hspace{2pt}}l@{\hspace{2pt}}l@{}}
\textit{Güneş} & \textit{gör-me-yen} & \textit{petunya-lar} & \textit{hemen} & \textit{[\underline{ölür}/*\underline{öler}].} \\
sun & see{\tiny \textsc{-neg-part}} & petunia{\tiny \textsc{-pl}} & immediately & dies\\ 
\end{tabular}
& \textit{Petunias that do not see the sun die immediately.}
\\
{\scriptsize Island Effects}       & \renewcommand{\arraystretch}{1.0}\begin{tabular}[t]{@{}l@{\hspace{2pt}}l@{\hspace{2pt}}l@{\hspace{2pt}}l@{\hspace{2pt}}l@{\hspace{2pt}}l@{\hspace{2pt}}l@{\hspace{2pt}}l@{}}
\textit{[\underline{Neyi}} & \textit{/*\underline{Onu}} & \textit{\underline{neden}]} & \textit{dükkan-a} & \textit{getir-en} & \textit{eleman} & \textit{azar} & \textit{işit-ti?}\\
{[}what & /*it & why{]} & shop{\tiny \textsc{-dat}} & bring{\tiny \textsc{-part}} & worker & scolding & hear{\tiny \textsc{-pst}}\\ 
\end{tabular}
& \textit{The worker who brought what to the store was scolded?}
\\ 
{\scriptsize Nominalization}    & \renewcommand{\arraystretch}{1.0}\begin{tabular}[t]{@{}l@{\hspace{2pt}}l@{\hspace{2pt}}l@{\hspace{2pt}}l@{\hspace{2pt}}l@{\hspace{2pt}}l@{\hspace{2pt}}l@{}}
\textit{Konu-nun} & \textit{tekrar} & \textit{[\underline{tartış-ıl-ma-sın-ı}} & \textit{/*\underline{tartış-ıl-dığ-ın-ı}]} & öner-iyor-um. \\
matter{\tiny \textsc{-gen}} & again & [discuss{\tiny \textsc{-pass-MA-poss-acc}} & /*discuss{\tiny \textsc{-pass-DIK-poss-acc}}] & suggest{\tiny \textsc{-prog-1sg}} \\ 
\end{tabular}
& \textit{I suggest that the matter be discussed again.}
\\ 
{\scriptsize NPI Licensing}        & \renewcommand{\arraystretch}{1.0}\begin{tabular}[t]{@{}l@{\hspace{2pt}}l@{\hspace{2pt}}l@{\hspace{2pt}}l@{\hspace{2pt}}l@{\hspace{2pt}}l@{\hspace{2pt}}l@{\hspace{2pt}}l@{}}
\textit{Kalabalığ-ın} & \textit{ön-ün-de} & \textit{[$\varnothing$} & \textit{/*\underline{hiç}]} & \textit{şarkı} & \textit{söyle-di-m.} \\
crowd{\tiny \textsc{-gen}} & front{\tiny \textsc{-poss-loc}} & {[}$\varnothing$ & /*ever{]} & song & sing{\tiny \textsc{-pst-1sg}} \\ 
\end{tabular}
& \textit{I (*ever) sang in front of a crowd.}
\\ 
{\scriptsize Passives}             & \renewcommand{\arraystretch}{1.0}\begin{tabular}[t]{@{}l@{\hspace{2pt}}l@{\hspace{2pt}}l@{\hspace{2pt}}l@{\hspace{2pt}}l@{\hspace{2pt}}l@{\hspace{2pt}}l@{\hspace{2pt}}l@{\hspace{2pt}}l@{}}
\textit{Sabah} & \textit{[$\varnothing$} & \textit{/*\underline{öğrenciler}} & \textit{\underline{tarafından}]} & \textit{okul} & \textit{bahçe-sin-de} & \textit{koş-ul-du.} \\
morning & {[}$\varnothing$ & /*student{\tiny \textsc{-pl}} & by{]} & school & yard{\tiny \textsc{-3sg.poss-loc}} & run{\tiny \textsc{-pass-pst}} \\ 
\end{tabular}
& \textit{$\sim$In the morning, it was ran in the school yard (*by the students).}
\\ 
{\scriptsize Quantifiers}          & \renewcommand{\arraystretch}{1.0}\begin{tabular}[t]{@{}l@{\hspace{2pt}}l@{\hspace{2pt}}l@{\hspace{2pt}}l@{\hspace{2pt}}l@{\hspace{2pt}}l@{\hspace{2pt}}l@{}}
\textit{Mağaza-da} & \textit{[$\varnothing$} & \textit{/*\underline{çoğu}]} & \textit{ayakkabı} & \textit{dene-di-m.} \\
store{\tiny \textsc{-loc}} & {[}$\varnothing$ & /*çoğu{]} & shoe & try\_on{\tiny \textsc{-pst-1sg}} \\ 
\end{tabular}
& \textit{I tried on shoes in the store.}
\\ 
{\scriptsize Relative Clauses}  & \renewcommand{\arraystretch}{1.0}\begin{tabular}[t]{@{}l@{\hspace{2pt}}l@{\hspace{2pt}}l@{\hspace{2pt}}l@{\hspace{2pt}}l@{\hspace{2pt}}l@{\hspace{2pt}}l@{\hspace{2pt}}l@{\hspace{2pt}}l@{}}
\textit{Sınav-da} & \textit{[\underline{gözetmen-in}} & \textit{/*\underline{gözetmen-i}]} & \textit{uyar-dığ-ı} & \textit{öğrenci} & \textit{yer-in-e} & \textit{geç-ti.} \\
exam{\tiny \textsc{-loc}} & {[}proctor{\tiny \textsc{-3sg.gen}} & /*proctor{\tiny \textsc{-acc}}{]} & warn{\tiny \textsc{-part-3sg.poss}} & student & place{\tiny \textsc{-poss-dat}} & move{\tiny \textsc{-pst}}\\ 
\end{tabular}
& \textit{The student whom the proctor quietly warned during the exam took his/her seat.}
\\ 
{\scriptsize Scrambling}           & \renewcommand{\arraystretch}{1.0}\begin{tabular}[t]{@{}l@{\hspace{2pt}}l@{\hspace{2pt}}l@{\hspace{2pt}}l@{\hspace{2pt}}l@{\hspace{2pt}}l@{\hspace{2pt}}l@{\hspace{2pt}}l@{}}
\textit{Hasan'ın} & \textit{[\underline{makale-yi}} & \textit{\underline{yaz-dığ-ın-ı}} & \textit{/*\underline{yaz-dığ-ın-ı}} & & \textit{\underline{makale-yi}]} & \textit{bil-iyor-um.} \\
Hasan{\tiny \textsc{-3sg.gen}} & article{\tiny \textsc{-acc}} & write{\tiny \textsc{-nmlz-3sg.poss-acc}} &  &  & & know{\tiny \textsc{-prog-1sg}} \\ 
\end{tabular}
& \textit{I know that Hasan wrote the article.}
\\ 
{\scriptsize Subject Agreement}    & \renewcommand{\arraystretch}{1.0}\begin{tabular}[t]{@{}l@{\hspace{2pt}}l@{\hspace{2pt}}l@{\hspace{2pt}}l@{\hspace{2pt}}l@{\hspace{2pt}}l@{\hspace{2pt}}l@{\hspace{2pt}}l@{\hspace{2pt}}l@{}}
\textit{[\underline{Doktor-lar}} & \textit{/*\underline{Doktor}]} & \textit{bu} & \textit{şart-ta} & \textit{çalış-mak} & \textit{zorunda} & \textit{değil-ler.} \\
{[}doctor{\tiny \textsc{-pl}} & /*doctor{]} & this & condition{\tiny \textsc{-loc}} & work{\tiny \textsc{-nmlz}} & obliged & {\tiny \textsc{neg-3pl}} \\ 
\end{tabular}
& \textit{[Doctors/*Doctor] do not have to work under these conditions.}
\\ 
{\scriptsize Suspended Affixation} & \renewcommand{\arraystretch}{1.0}\begin{tabular}[t]{@{}l@{\hspace{2pt}}l@{\hspace{2pt}}l@{\hspace{2pt}}l@{\hspace{2pt}}l@{\hspace{2pt}}l@{\hspace{2pt}}l@{\hspace{2pt}}l@{\hspace{2pt}}l@{\hspace{2pt}}l@{}}
\textit{Akşam} & \textit{kız-lar-la} & \textit{parti-ye} & \textit{[\underline{git-ti-k}} & \textit{/*\underline{git}]} & \textit{ve} & \textit{çok} & \textit{eğlen-di-k.}  \\
evening & girl{\tiny \textsc{-pl-com}} & party{\tiny \textsc{-dat}} & {[}go{\tiny \textsc{-pst-1pl}} & /*go{]} & and & very & have\_fun{\tiny \textsc{-pst-1pl}}  \\ 
\end{tabular}
& \textit{In the evening, we went to a party with the girls and had a lot of fun.}            
\\ \bottomrule
\end{tabular}
\caption{Glossed minimal pairs for each phenomenon in TurBLiMP. The differences are underlined.}
\label{tab:trblimp}
\end{table*}

\subsection{Phenomena}

We consider 16 different grammatical phenomena, some of which are cross-lingually present in other benchmarks, alongside a few language-specific ones such as suspended affixation (see Table~\ref{tab:trblimp} for a complete overview with examples).

\paragraph{\textsc{Anaphor Agreement}} The anaphoric reflexive pronoun \textit{kendi} agrees with its referent through number and person inflections. Unacceptable sentences in this category feature inflected forms of \textit{kendi} with incorrect agreement.

\paragraph{\textsc{Argument Structure (Transitive)}} Turkish has a nominative-accusative case marking system where the direct object of a sentence is marked by the accusative case. However, a special subset of verbs assigns lexical case to their objects, deviating from structural case assignment. Unacceptable sentences feature objects with incorrect case endings, such as dative.

\paragraph{\textsc{Argument Structure (Ditransitive)}} The prototypical Turkish ditransitive construction applies a dative case marker to the indirect object. However, verbs assigning lexical case can deviate from the general trend. Here too, unacceptable sentences feature objects with incorrect case endings.

\paragraph{\textsc{Binding}} Principle B in Binding Theory \citep{chomsky1981} asserts that pronouns should be free in their binding domain, implying that pronouns should not refer to another entity in the same immediate clause. 
Unacceptable sentences are created by swapping an anaphora coreferring with the subject with a pronoun of similar features.

\paragraph{\textsc{Determiners}} While determiners are largely optional in Turkish, the indefinite article \textit{bir} is sometimes required. 
When a direct object occurs immediately before the verb, its accusative case ending can be omitted.
If such an object is modified by a relative clause, the indefinite article must precede the noun head \citep{arslan2009referentiality}. Unacceptable sentences in this phenomenon omit the obligatory determiner.

\paragraph{\textsc{Ellipsis}} This phenomenon deals with a specific type of ellipsis called backward gapping. For coordinated clauses in Turkish, it is possible to omit the verb in the first clause, leading to a gap which is resolved by the verb in the second clause. Turkish only licenses this if both clauses maintain parallel word order \citep{Bozsahin2000}. Acceptable sentences show the same subject-object order across clauses while unacceptable ones alternate their order.

\paragraph{\textsc{Irregular Forms}} The aorist is an aspect/ mood marker with three allomorphs -r, -Ir (high vowel harmony), and -Ar (non-high vowel harmony). While monosyllabic verbs take -Ar, a specific subset of irregular verbs take -Ir \citep{NAKIPOĞLU_UZUNDAĞ_KETREZ_2023}. Unacceptable sentences feature an incorrect -Ar form.

\paragraph{\textsc{Island Effects}} We focus on a specific type of island constraint in which complex noun phrases are modified by a relative clause containing a wh-phrase. 
The occurrence of the wh-phrase is only permitted if the wh-phrase is \textit{not} an adjunct \citep{sinancakir2016}.  Acceptable sentences contain argument wh-phrases like who or what, while unacceptable ones contain wh-adjuncts such as how or why.

\paragraph{\textsc{Nominalization}}
Turkish extensively uses a derivational process called nominalization, where verbal bases take suffixes (like -DIK, \mbox{-mA}, and others) to form noun phrases. A category of Turkish verbs only selects complement clauses with -DIK, while others only allow -mA \citep{Kornfilt+2003+129+216}. Correspondingly, minimal pairs contain verbs with the correct and incorrect nominalization suffixes.

\paragraph{\textsc{NPI Licensing}} This phenomenon deals with Turkish negative polarity items such as \textit{hiç}, \textit{kimse}, \textit{hiçbir}, \textit{hiçbir şey}, and \textit{asla}. NPIs occur in contexts where the predicate is negated. Acceptable sentences either omit the NPI or use placeholder indefinite pronouns, while unacceptable ones feature an NPI with a predicate that is not negated. 

\paragraph{\textsc{Passives}} Turkish licenses the passivization of intransitive verbs via passive suffixes, creating impersonal (vs. personal) passives. While personal passives permit optional by-phrases to express agents, impersonal passives prohibit them \citep{ozsoy2009}. Thus, acceptable sentences omit by-phrases, while unacceptable ones include them.

\paragraph{\textsc{Quantifiers}} Turkish quantifiers such as \textit{her} and \textit{çoğu} can only occur with accusative-marked nouns \citep{a5cf88a7-e671-314d-be95-4caaea877f88}. All minimal pairs for this phenomenon feature direct objects without accusative marking. Unacceptable sentences include a quantifier before the bare noun while acceptable sentences omit it.

\paragraph{\textsc{Relative Clauses}} 
Turkish uses participle suffixes -DIK and -An to form object and subject relative clauses \citep{göksel2005turkish}. \mbox{-DIK} clauses feature genitive-possessive agreement. The subject takes genitive case and the verb carries possessive agreement. In subject relative clauses with -An, only the object (if present) is case-marked. Minimal pairs target an argument preceding the nominalized verb. Acceptability depends on whether this noun is inflected with a genitive or non-genitive case ending.

\paragraph{\textsc{Scrambling}} Turkish shows word order flexibility and allows postverbal scrambling. This means that constituents can appear after the verb in certain contexts. However, local postverbal scrambling from an embedded clause is prohibited \citep{kornfiltscrambling}. Acceptable sentences position the object before the embedded verb while unacceptable sentences feature them in the opposite order.

\paragraph{\textsc{Subject Agreement}} Turkish realizes subject-verb agreement via person/number suffixes. Gender agreement is absent. A notable feature is third-person syncretism. The same verb inflection can indicate either a third-person singular or plural subject. However, a plural-inflected verb cannot co-occur with a singular subject. Unacceptable sentences either involve singular subjects with plural verbs or pronoun mismatches with first/second-person agreement.

\paragraph{\textsc{Suspended Affixation}} Suspended affixation refers to a phenomenon where a shared suffix applies to all conjuncts in a coordinated structure, rather than being repeated. Turkish does not allow suspended affixation for predicates inflected only with the past tense suffix -DI \citep{kutay2019}. Minimal pairs feature two coordinated past-tense clauses. Acceptable sentences inflect both verbs, while unacceptable ones omit inflection on the first.

\subsection{Benchmark Creation}

In the creation of TurBLiMP, we opted for the more labor-intensive process of manually crafting sentences. 10 initial samples per each phenomenon were created entirely manually to establish clear guidelines. This first step ensured that each pair differed only minimally while accurately capturing the targeted grammatical contrasts.

\paragraph{Semi-automatic augmentations}
To enhance lexical diversity, we then adopted a semi-automated workflow in which a masked Turkish LM, BERTurk \citep{stefan_schweter_2020_3770924} is used to suggest lexical replacements at random positions of each manually created sentence. We verified and adjusted each replacement manually to ensure acceptability. This process yielded 100 samples per phenomenon. In a final fully-automated augmentation step, BERTurk was used to generate a list of contextually appropriate words for replacement (e.g. \textit{woman} or \textit{boy} for \textit{girl}). We use the Turkish morphology pipeline by \citet{akin2007zemberek} to inflect them with the same morphological features when applicable. At the end of this process, our 100 manually validated pairs increase to 1000 pairs per phenomenon. Our three-fold approach balanced scalability with linguistic precision, resulting in a robust benchmark for evaluating Turkish LMs.

\subsection{Experimental Paradigms}
We further assess the robustness of LMs' syntactic abilities by focusing on two salient properties of Turkish: (i) word order flexibility and (ii) subordination through morphological processes, both discussed in Section~\ref{sect:turkish-lm}. Word order variations provide a useful framework for testing the effect of word order biases on syntactic competence, extending the types of variations covered by the existing minimal pair benchmarks \cite{linzen-etal-2016-assessing,mueller-etal-2020-cross}.
Subordination is a particularly interesting case to study the interplay between syntactic competence and morphological generalization, broadening the scope of current word-level evaluations \cite{ismayilzada-etal-2025-evaluating}.

We generate word order and subordinating variations for two of the TurBLiMP phenomena (Transitive and Ditransitive Argument Structure) chosen for their flexibility for manipulation. We derive all 6 subject/verb/object orders and 4 different subordination structures for each minimal pair.
Complete examples of experimental paradigms and details about how they were created
are provided in Appendix \ref{sec:appendix}. The experimental paradigms add a total of 2,000 minimal pairs to the 16,000 pairs forming the base TurBLiMP, and considerably extend our benchmark's utility for investigating controlled linguistic variations.

\section{Human Acceptability Judgments}
To validate our benchmark, we collected acceptability judgments from 30 native Turkish speakers using a 7-point Likert scale (1: completely unacceptable, 7: completely acceptable). While previous BLiMP variants rely on forced-choice tasks for data validation, BLiMP-NL \citep{10.1162/coli_a_00559} collects Likert scale responses to capture the gradient nature of acceptability judgments. We followed their approach to provide a benchmark that allows for fine-grained evaluation of model-human alignment. Our participant pool was mixed, comprising 17 linguistics students and 13 non-linguists. The study was carried out via an anonymous online survey. Appendix \ref{sec:appendix_survey} includes a screenshot of survey instructions. Each participant rated 216 sentences spanning 16 linguistic phenomena as well as 20 experimental paradigms. 3 acceptable and 3 unacceptable sentences were included for each grammatical category, and the acceptability conditions were flipped between the two survey versions. 

\begin{figure}[ht]
  \includegraphics[width=0.48\textwidth]{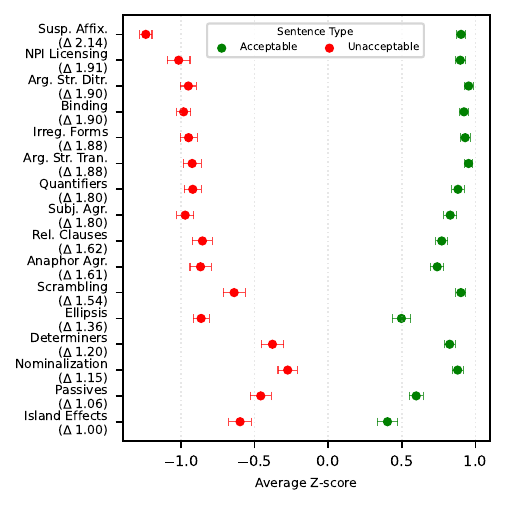}
  \caption{Mean acceptability judgments for 16 TurBLiMP phenomena. Likert scale ratings are transformed to z-scores. Error bars show standard errors of the mean.}
  \label{fig:acceptability_base}
\end{figure}

Figure \ref{fig:acceptability_base} reports average acceptability judgments for each phenomenon. Additional participant rating statistics are provided in Appendix \ref{sec:appendix_prop}. The responses are first normalized by transforming Likert scores to z-scores. We assume that every participant uses the scale slightly differently, and this step ensures comparability across participants.

Overall, our analysis reveals that participants made clear distinctions between acceptable and unacceptable sentences. The clear separation in z-scores between grammatical and ungrammatical constructions confirms that the targeted syntactic distinctions are perceptible to native speakers. We can also note that some phenomena such as Island Effects, Passives, and Nominalization were less discriminable than others. 

\begin{table*}[ht]
\centering
\small
\adjustbox{max width=\textwidth}{%
\renewcommand{\arraystretch}{1}
\setlength{\tabcolsep}{3pt}
\hspace*{-0.47em}\begin{tabular}{@{}lrrrrrrrrrrrrrr@{ }}
\multicolumn{1}{@{}l}{Phenomenon} & \multicolumn{1}{l}{\makebox[0pt][l]{\begin{turn}{50}Goldfish 5MB\end{turn}}} & \multicolumn{1}{l}{\makebox[0pt][l]{\begin{turn}{50}Goldfish 10MB\end{turn}}} & \multicolumn{1}{l}{\makebox[0pt][l]{\begin{turn}{50}Goldfish 100MB\end{turn}}} & \multicolumn{1}{l}{\makebox[0pt][l]{\begin{turn}{50}Goldfish 1000MB\end{turn}}} & \multicolumn{1}{l}{\makebox[0pt][l]{\begin{turn}{50}BERTurk\end{turn}}} & \multicolumn{1}{l}{\makebox[0pt][l]{\begin{turn}{50}cosmosGPT\end{turn}}} & \multicolumn{1}{l}{\makebox[0pt][l]{\begin{turn}{50}Gemma 3\end{turn}}} & \multicolumn{1}{l}{\makebox[0pt][l]{\begin{turn}{50}Qwen 2.5\end{turn}}} & \multicolumn{1}{l}{\makebox[0pt][l]{\begin{turn}{50}Llama 3.1\end{turn}}} & \multicolumn{1}{l}{\makebox[0pt][l]{\begin{turn}{50}Aya Expanse\end{turn}}} & \multicolumn{1}{l}{\makebox[0pt][l]{\begin{turn}{50}Gemma 2\end{turn}}} & \multicolumn{1}{l}{\makebox[0pt][l]{\begin{turn}{50}EuroLLM\end{turn}}} & \multicolumn{1}{l}{\makebox[0pt][l]{\begin{turn}{50}Gemma 3\end{turn}}} \\
\toprule
Anaphor Agreement & \cellcolor[rgb]{1,0.84,0.66} 42.0 & \cellcolor[rgb]{1,0.89,0.69} 44.4 & \cellcolor[rgb]{0.76,0.88,0.63} 69.9 & \cellcolor[rgb]{0.50,0.76,0.46} 90.8 & \cellcolor[rgb]{0.42,0.72,0.41} 97.7 & \cellcolor[rgb]{0.47,0.75,0.45} 93.0 & \cellcolor[rgb]{0.49,0.75,0.45} 92.1 & \cellcolor[rgb]{0.60,0.80,0.52} 83.2 & \cellcolor[rgb]{0.52,0.76,0.47} 89.5 & \cellcolor[rgb]{0.53,0.77,0.48} 88.5 & \cellcolor[rgb]{0.51,0.76,0.47} 89.8 & \cellcolor[rgb]{0.46,0.74,0.44} 94.1 & \cellcolor[rgb]{0.47,0.75,0.44} 93.2 \\
Argument Str. Tran. & \cellcolor[rgb]{0.93,0.96,0.73} 56.1 & \cellcolor[rgb]{0.99,1.00,0.78} 50.6 & \cellcolor[rgb]{0.54,0.78,0.49} 87.9 & \cellcolor[rgb]{0.41,0.71,0.40} 98.3 & \cellcolor[rgb]{0.40,0.71,0.40} 99.1 & \cellcolor[rgb]{0.40,0.71,0.40} 99.3 & \cellcolor[rgb]{0.43,0.73,0.42} 96.5 & \cellcolor[rgb]{0.60,0.81,0.53} 82.5 & \cellcolor[rgb]{0.49,0.76,0.46} 91.3 & \cellcolor[rgb]{0.48,0.75,0.45} 92.8 & \cellcolor[rgb]{0.49,0.75,0.45} 92.2 & \cellcolor[rgb]{0.42,0.72,0.41} 97.6 & \cellcolor[rgb]{0.40,0.71,0.40} 99.1 \\
Argument Str. Ditr. & \cellcolor[rgb]{0.81,0.91,0.66} 65.3 & \cellcolor[rgb]{0.96,0.98,0.76} 53.2 & \cellcolor[rgb]{0.60,0.81,0.53} 82.5 & \cellcolor[rgb]{0.48,0.75,0.45} 92.3 & \cellcolor[rgb]{0.44,0.73,0.42} 96.1 & \cellcolor[rgb]{0.42,0.72,0.41} 98.0 & \cellcolor[rgb]{0.43,0.73,0.42} 96.6 & \cellcolor[rgb]{0.49,0.75,0.45} 91.7 & \cellcolor[rgb]{0.52,0.76,0.47} 89.6 & \cellcolor[rgb]{0.45,0.74,0.43} 94.6 & \cellcolor[rgb]{0.43,0.72,0.42} 96.8 & \cellcolor[rgb]{0.43,0.72,0.42} 96.7 & \cellcolor[rgb]{0.42,0.72,0.41} 97.6 \\
Binding & \cellcolor[rgb]{1,0.42,0.33} 21.1 & \cellcolor[rgb]{1,0.51,0.40} 25.8 & \cellcolor[rgb]{0.84,0.92,0.68} 63.3 & \cellcolor[rgb]{0.48,0.75,0.45} 92.3 & \cellcolor[rgb]{0.40,0.71,0.40} 99.0 & \cellcolor[rgb]{0.40,0.71,0.40} 99.2 & \cellcolor[rgb]{0.43,0.73,0.42} 96.5 & \cellcolor[rgb]{0.50,0.76,0.46} 91.0 & \cellcolor[rgb]{0.42,0.72,0.41} 97.1 & \cellcolor[rgb]{0.44,0.73,0.42} 95.6 & \cellcolor[rgb]{0.42,0.72,0.41} 97.9 & \cellcolor[rgb]{0.41,0.71,0.40} 98.6 & \cellcolor[rgb]{0.41,0.71,0.40} 98.2 \\
Determiners & \cellcolor[rgb]{1,0.36,0.28} 18.1 & \cellcolor[rgb]{1,0.52,0.40} 25.9 & \cellcolor[rgb]{0.73,0.87,0.61} 72.5 & \cellcolor[rgb]{0.46,0.74,0.44} 94.3 & \cellcolor[rgb]{0.40,0.71,0.40} 99.3 & \cellcolor[rgb]{0.46,0.74,0.44} 94.2 & \cellcolor[rgb]{0.54,0.78,0.49} 87.9 & \cellcolor[rgb]{0.69,0.85,0.58} 75.4 & \cellcolor[rgb]{0.63,0.82,0.55} 80.1 & \cellcolor[rgb]{0.55,0.78,0.49} 86.7 & \cellcolor[rgb]{0.49,0.75,0.45} 91.7 & \cellcolor[rgb]{0.47,0.75,0.44} 93.3 & \cellcolor[rgb]{0.44,0.73,0.42} 96.1 \\
Ellipsis & \cellcolor[rgb]{1,0.66,0.52} 33.0 & \cellcolor[rgb]{1,0.60,0.47} 30.2 & \cellcolor[rgb]{0.78,0.89,0.64} 68.0 & \cellcolor[rgb]{1,0.29,0.23} 14.9 & \cellcolor[rgb]{0.54,0.78,0.49} 87.5 & \cellcolor[rgb]{1,0.80,0.63} 40.2 & \cellcolor[rgb]{0.87,0.94,0.70} 60.5 & \cellcolor[rgb]{1,0.87,0.68} 43.6 & \cellcolor[rgb]{0.83,0.92,0.67} 63.8 & \cellcolor[rgb]{0.84,0.92,0.68} 62.7 & \cellcolor[rgb]{0.91,0.95,0.73} 57.5 & \cellcolor[rgb]{0.72,0.86,0.60} 73.2 & \cellcolor[rgb]{0.75,0.88,0.62} 70.4 \\
Irregular Forms & \cellcolor[rgb]{0.39,0.71,0.39} 100.0 & \cellcolor[rgb]{0.39,0.71,0.39} 100.0 & \cellcolor[rgb]{0.39,0.71,0.39} 100.0 & \cellcolor[rgb]{0.39,0.71,0.39} 100.0 & \cellcolor[rgb]{0.40,0.71,0.39} 99.6 & \cellcolor[rgb]{0.39,0.71,0.39} 100.0 & \cellcolor[rgb]{0.39,0.71,0.39} 100.0 & \cellcolor[rgb]{0.39,0.71,0.39} 100.0 & \cellcolor[rgb]{0.39,0.71,0.39} 100.0 & \cellcolor[rgb]{0.40,0.71,0.40} 99.4 & \cellcolor[rgb]{0.39,0.71,0.39} 99.9 & \cellcolor[rgb]{0.39,0.71,0.39} 100.0 & \cellcolor[rgb]{0.39,0.71,0.39} 100.0 \\
Island Effects & \cellcolor[rgb]{0.47,0.74,0.44} 93.7 & \cellcolor[rgb]{0.48,0.75,0.45} 92.8 & \cellcolor[rgb]{0.62,0.82,0.54} 81.0 & \cellcolor[rgb]{0.66,0.84,0.56} 78.0 & \cellcolor[rgb]{0.98,0.99,0.77} 51.2 & \cellcolor[rgb]{0.51,0.76,0.47} 89.8 & \cellcolor[rgb]{0.69,0.85,0.58} 75.4 & \cellcolor[rgb]{0.56,0.79,0.50} 86.0 & \cellcolor[rgb]{0.65,0.83,0.56} 78.3 & \cellcolor[rgb]{0.73,0.87,0.61} 72.5 & \cellcolor[rgb]{0.69,0.85,0.58} 75.8 & \cellcolor[rgb]{0.64,0.83,0.55} 79.3 & \cellcolor[rgb]{0.69,0.85,0.58} 75.5 \\
Nominalization & \cellcolor[rgb]{0.80,0.90,0.65} 66.4 & \cellcolor[rgb]{0.71,0.86,0.60} 73.9 & \cellcolor[rgb]{0.50,0.76,0.46} 90.8 & \cellcolor[rgb]{0.47,0.75,0.44} 93.3 & \cellcolor[rgb]{0.42,0.72,0.41} 97.4 & \cellcolor[rgb]{0.43,0.72,0.42} 97.0 & \cellcolor[rgb]{0.45,0.73,0.43} 94.9 & \cellcolor[rgb]{0.46,0.74,0.44} 94.0 & \cellcolor[rgb]{0.45,0.73,0.43} 95.3 & \cellcolor[rgb]{0.48,0.75,0.45} 92.4 & \cellcolor[rgb]{0.44,0.73,0.42} 95.9 & \cellcolor[rgb]{0.45,0.73,0.43} 95.2 & \cellcolor[rgb]{0.43,0.73,0.42} 96.6 \\
NPI Licensing & \cellcolor[rgb]{0.47,0.74,0.44} 93.4 & \cellcolor[rgb]{0.46,0.74,0.44} 94.3 & \cellcolor[rgb]{0.49,0.75,0.45} 91.9 & \cellcolor[rgb]{0.41,0.71,0.40} 98.1 & \cellcolor[rgb]{0.45,0.73,0.43} 95.0 & \cellcolor[rgb]{0.41,0.71,0.40} 98.5 & \cellcolor[rgb]{0.42,0.72,0.41} 97.7 & \cellcolor[rgb]{0.46,0.74,0.44} 94.0 & \cellcolor[rgb]{0.45,0.73,0.43} 95.4 & \cellcolor[rgb]{0.43,0.72,0.42} 96.9 & \cellcolor[rgb]{0.42,0.72,0.41} 97.1 & \cellcolor[rgb]{0.43,0.72,0.42} 96.9 & \cellcolor[rgb]{0.41,0.71,0.40} 98.1 \\
Passives & \cellcolor[rgb]{0.39,0.71,0.39} 100.0 & \cellcolor[rgb]{0.39,0.71,0.39} 100.0 & \cellcolor[rgb]{0.39,0.71,0.39} 99.9 & \cellcolor[rgb]{0.39,0.71,0.39} 99.9 & \cellcolor[rgb]{0.62,0.82,0.54} 81.3 & \cellcolor[rgb]{0.40,0.71,0.40} 99.4 & \cellcolor[rgb]{0.39,0.71,0.39} 100.0 & \cellcolor[rgb]{0.39,0.71,0.39} 100.0 & \cellcolor[rgb]{0.40,0.71,0.39} 99.6 & \cellcolor[rgb]{0.40,0.71,0.40} 98.8 & \cellcolor[rgb]{0.40,0.71,0.40} 99.5 & \cellcolor[rgb]{0.39,0.71,0.39} 99.7 & \cellcolor[rgb]{0.39,0.71,0.39} 99.9 \\
Quantifiers & \cellcolor[rgb]{0.40,0.71,0.40} 99.0 & \cellcolor[rgb]{0.40,0.71,0.40} 99.0 & \cellcolor[rgb]{0.40,0.71,0.40} 99.0 & \cellcolor[rgb]{0.40,0.71,0.40} 98.9 & \cellcolor[rgb]{0.41,0.71,0.40} 98.4 & \cellcolor[rgb]{0.40,0.71,0.40} 99.0 & \cellcolor[rgb]{0.40,0.71,0.40} 99.0 & \cellcolor[rgb]{0.41,0.71,0.40} 98.4 & \cellcolor[rgb]{0.41,0.71,0.40} 98.5 & \cellcolor[rgb]{0.42,0.72,0.41} 98.0 & \cellcolor[rgb]{0.41,0.71,0.40} 98.7 & \cellcolor[rgb]{0.40,0.71,0.40} 98.9 & \cellcolor[rgb]{0.40,0.71,0.40} 99.0 \\
Relative Clauses & \cellcolor[rgb]{1,0.96,0.76} 48.3 & \cellcolor[rgb]{1,0.98,0.77} 49.4 & \cellcolor[rgb]{0.67,0.84,0.57} 76.7 & \cellcolor[rgb]{0.61,0.81,0.53} 82.0 & \cellcolor[rgb]{0.41,0.71,0.40} 98.5 & \cellcolor[rgb]{0.47,0.75,0.45} 93.0 & \cellcolor[rgb]{0.60,0.81,0.53} 82.4 & \cellcolor[rgb]{0.74,0.87,0.62} 71.2 & \cellcolor[rgb]{0.63,0.82,0.55} 80.3 & \cellcolor[rgb]{0.63,0.82,0.54} 80.6 & \cellcolor[rgb]{0.62,0.82,0.54} 80.7 & \cellcolor[rgb]{0.58,0.80,0.52} 83.9 & \cellcolor[rgb]{0.62,0.81,0.54} 81.5 \\
Scrambling & \cellcolor[rgb]{0.71,0.85,0.59} 74.1 & \cellcolor[rgb]{0.55,0.78,0.49} 86.6 & \cellcolor[rgb]{0.40,0.71,0.39} 99.6 & \cellcolor[rgb]{0.39,0.71,0.39} 99.9 & \cellcolor[rgb]{0.39,0.71,0.39} 100.0 & \cellcolor[rgb]{0.39,0.71,0.39} 99.7 & \cellcolor[rgb]{0.40,0.71,0.40} 99.0 & \cellcolor[rgb]{0.39,0.71,0.39} 99.9 & \cellcolor[rgb]{0.41,0.71,0.40} 98.2 & \cellcolor[rgb]{0.40,0.71,0.40} 98.8 & \cellcolor[rgb]{0.39,0.71,0.39} 100.0 & \cellcolor[rgb]{0.39,0.71,0.39} 100.0 & \cellcolor[rgb]{0.39,0.71,0.39} 100.0 \\
Subject Agreement & \cellcolor[rgb]{1,0.89,0.69} 44.4 & \cellcolor[rgb]{1,0.83,0.65} 41.5 & \cellcolor[rgb]{0.58,0.80,0.51} 84.1 & \cellcolor[rgb]{0.45,0.73,0.43} 94.8 & \cellcolor[rgb]{0.40,0.71,0.40} 98.8 & \cellcolor[rgb]{0.42,0.72,0.41} 97.5 & \cellcolor[rgb]{0.52,0.77,0.47} 89.1 & \cellcolor[rgb]{0.60,0.80,0.53} 82.8 & \cellcolor[rgb]{0.58,0.79,0.51} 84.8 & \cellcolor[rgb]{0.49,0.75,0.45} 91.7 & \cellcolor[rgb]{0.51,0.76,0.46} 90.6 & \cellcolor[rgb]{0.47,0.74,0.44} 93.7 & \cellcolor[rgb]{0.48,0.75,0.45} 92.7 \\
Suspended Affixation & \cellcolor[rgb]{0.91,0.96,0.73} 57.2 & \cellcolor[rgb]{0.82,0.91,0.67} 64.8 & \cellcolor[rgb]{0.47,0.74,0.44} 93.8 & \cellcolor[rgb]{0.41,0.71,0.40} 98.3 & \cellcolor[rgb]{0.39,0.71,0.39} 100.0 & \cellcolor[rgb]{0.40,0.71,0.39} 99.6 & \cellcolor[rgb]{0.39,0.71,0.39} 99.8 & \cellcolor[rgb]{0.42,0.72,0.41} 97.5 & \cellcolor[rgb]{0.41,0.71,0.40} 98.2 & \cellcolor[rgb]{0.40,0.71,0.40} 98.9 & \cellcolor[rgb]{0.40,0.71,0.40} 99.5 & \cellcolor[rgb]{0.39,0.71,0.39} 99.7 & \cellcolor[rgb]{0.39,0.71,0.39} 100.0 \\
\midrule
Model Average & \cellcolor[rgb]{0.84,0.92,0.68} 63.3 & \cellcolor[rgb]{0.82,0.91,0.67} 64.5 & \cellcolor[rgb]{0.57,0.79,0.51} 85.1 & \cellcolor[rgb]{0.52,0.77,0.47} 89.1 & \cellcolor[rgb]{0.47,0.74,0.44} 93.7 & \cellcolor[rgb]{0.47,0.74,0.44} 93.6 & \cellcolor[rgb]{0.49,0.75,0.45} 91.7 & \cellcolor[rgb]{0.55,0.78,0.49} 87.0 & \cellcolor[rgb]{0.51,0.76,0.47} 90.0 & \cellcolor[rgb]{0.51,0.76,0.46} 90.6 & \cellcolor[rgb]{0.49,0.75,0.46} 91.5 & \cellcolor[rgb]{0.47,0.74,0.44} 93.8 & \cellcolor[rgb]{0.47,0.74,0.44} 93.6 \\
\addlinespace[0.17em]
Human Correlation & -0.30 & -0.30 & 0.01 & 0.16 & \textbf{0.65} & 0.25 & 0.16 & -0.07 & -0.01 & 0.25 & 0.09 & 0.17 & 0.17 \\
\bottomrule
\addlinespace[0.37em]
Parameter Count & \multicolumn{1}{c}{\footnotesize \texttt{39M}} & \multicolumn{1}{c}{\footnotesize \texttt{39M}} & \multicolumn{1}{c}{\footnotesize \texttt{125M}} & \multicolumn{1}{c}{\footnotesize \texttt{125M}} & \multicolumn{1}{c}{\footnotesize \texttt{185M}} & \multicolumn{1}{c}{\footnotesize \texttt{774M}} & \multicolumn{1}{c}{\footnotesize \texttt{4B}} & \multicolumn{1}{c}{\footnotesize \texttt{7B}} & \multicolumn{1}{c}{\footnotesize \texttt{8B}} & \multicolumn{1}{c}{\footnotesize \texttt{8B}} & \multicolumn{1}{c}{\footnotesize \texttt{9B}} & \multicolumn{1}{c}{\footnotesize \texttt{9B}} & \multicolumn{1}{c}{\footnotesize \texttt{12B}} \\
\cmidrule(lr){2-7} \cmidrule(lr){8-14}
Training Text & \multicolumn{6}{c}{\footnotesize Monolingual} & \multicolumn{7}{c}{\footnotesize Multilingual} \\
\end{tabular}}
\caption{Accuracy scores of each model across the linguistic phenomena in TurBLiMP. The red-green color gradient indicates performance, ranging from low to high. Significant Pearson correlations to the human judgments ($p < 0.05$) are indicated in boldface.}
\label{tab:heatmap}
\end{table*}

\section{Experimental Setup}

\paragraph{Monolingual models} We employed the Goldfish series \cite{chang-etal-2024-goldfish}, a series of causal LMs with fixed architecture trained on varying training data sizes (5MB, 10MB, 100MB, and 1000MB). Another monolingual model we used is BERTurk \citep{stefan_schweter_2020_3770924}, a 185M-parameter Turkish masked LM. With a vocabulary size of 128k, it is the only masked LM in our set of monolingual models. The largest monolingual model that we test is cosmosGPT \citep{kesgin2024introducing}, a 774M-parameter GPT-2-based model pretrained on Turkish web corpora and books.

\paragraph{Multilingual models} The evaluated multilingual models include Qwen 2.5 7B \citep{qwen2025qwen25technicalreport}, Llama 3.1 8B \citep{meta2024llama3.1}, Aya Expanse 8B \citep{dang2024ayaexpansecombiningresearch}, Gemma 2 7B \citep{gemmateam2024gemma2improvingopen}, Gemma 3 4B and 12B \citep{gemmateam2025gemma3technicalreport}, as well as EuroLLM 9B \citep{martins2024eurollmmultilinguallanguagemodels}.
For a balanced comparison between the various models, we employed comparable parameter sizes ranging from 4B to 12B. Notably, Aya Expanse is the only instruction-tuned variant in our set of multilingual models, supporting 23 languages including Turkish. The Gemma series also boast multilinguality with Gemma 3 providing support for over 140 languages. EuroLLM prioritizes the coverage of European languages alongside a few others including Turkish. 

As our evaluation metric for model performance, we computed entire-sequence log probabilities for acceptable and unacceptable sentences in each pair using the \texttt{minicons} library \cite{misra2022minicons, kauf2023better}. Accuracy scores reflect the proportion of pairs where the model assigned a higher probability to the acceptable sentence. We also report Pearson's correlation between human and model evaluations, calculated from the difference between average scores of acceptable and unacceptable sentences.

\section{Results}

Model performances across linguistic phenomena are summarized in Table \ref{tab:heatmap}. The results reveal that, more often than not, models were able to rate the acceptable sentence higher than its unacceptable counterpart. Some particular phenomena pose challenges for all the models. Ellipsis proved particularly difficult, with scores ranging from 14.9 to 87.5. Other challenging phenomena include Island Effects, Relative Clauses, and Determiners.

Island Effects, Determiners, and Ellipsis also happen to be some of the phenomena with the lowest mean rating difference in acceptability judgments collected from native speakers as seen in Figure \ref{fig:acceptability_base}. We should note that participants preserved a clear acceptability contrast with these phenomena as well. In the case of Ellipsis, considerably low model performances are not consistent with the collected judgments. Though Ellipsis and Scrambling both manipulate word order, models handle Scrambling well. Thus, Ellipsis scores cannot be attributed to general order-manipulation difficulty.

\begin{figure}[ht]
  \includegraphics[width=0.48\textwidth]{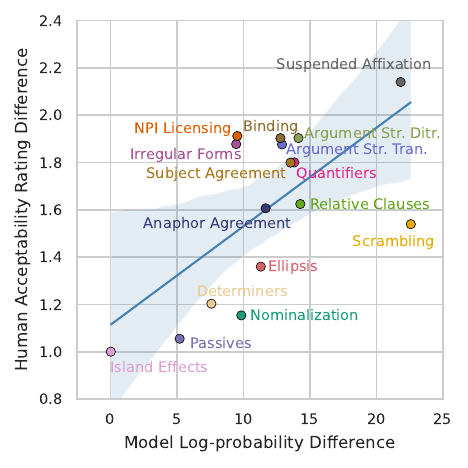}
  \caption{Correlation between the BERTurk model and human acceptability judgments across phenomena. (Pearson's $r = 0.65$, $p = 0.007$) Each data point corresponds to the average difference per phenomenon.}
  \label{fig:berturkcorr}
\end{figure}

We see that the monolingual models BERTurk and cosmosGPT tend to outperform their multilingual counterparts. Their performance is comparable to the best multilingual models EuroLLM and \mbox{Gemma 3 12B}. BERTurk is the only model that shows a strong cross-phenomenon correlation with human acceptability ratings, as illustrated in Figure \ref{fig:berturkcorr}. This is worth noting given that BERTurk is the only masked language model that we have tested. None of the other models had a statistically significant correlation in either direction. While the reported correlations are based on sequence log probabilities, we also experimented with using Syntactic Log-Odds Ratio (SLOR) \citep{pauls-klein-2012-large} scores to rule out the possibility that poor correlations were an artifact of the chosen metric. As an acceptability measure, SLOR normalizes sequence probabilities by controlling for sentence length and word frequency. SLOR-based correlations are provided in Appendix \ref{sec:slor}.

Multilingual models generally show better performance with increasing model sizes, but exceptions exist. Gemma 3 4B outperforms Gemma 2 8B, and EuroLLM 9B slightly surpasses Gemma 3 12B. The superior performance of EuroLLM 9B over the same-sized Gemma 2 9B may stem from better distribution of training data across languages.

Finally, the Goldfish model series reveals the effect of training data size on performance. Models with larger training data sizes typically achieve better performance, though some counter-intuitive patterns emerge near random-chance levels. While more data generally improves learning, this pattern does not hold when acceptable sentences are consistently shorter than unacceptable ones. In addition to the reported model results, a supplementary analysis of surface-level confounds such as sentence length and subword counts can be found in Appendix \ref{sec:phenomenon_agnostic}.

\subsection{Effect of Word Order}

Our word order paradigm results for the best monolingual (BERTurk) and multilingual (EuroLLM) models are illustrated in Table \ref{tab:word_order}. By manipulating minimal pairs for the Transitive and Ditransitive Argument Structure phenomena, we examine how different word orders affect performance.

\begin{table}[ht]
\centering
\small
\adjustbox{max width=0.48\textwidth}{%
\renewcommand{\arraystretch}{1.2}
\setlength{\tabcolsep}{3pt}
\begin{tabular}{lllrrrrrr}
\toprule
\multicolumn{1}{c}{} & Data & Metric & SOV & SVO & OSV & OVS & VSO & VOS \\
\midrule

\multirow{4}{*}{\rotatebox{90}{Human}} & 

\citeauthor{SLOBIN1982229} & Freq. & \cellcolor[rgb]{1.00,1.00,0.76} \textbf{48.0} & \cellcolor[rgb]{1.00,1.00,0.87} \textbf{25.0} & \cellcolor[rgb]{1.00,1.00,0.96} 8.0 & \cellcolor[rgb]{1.00,1.00,0.93} 13.0 & \cellcolor[rgb]{1.00,1.00,0.97} 6.0 & \cellcolor[rgb]{1.00,1.00,1.00} 0.0 \\
 & BOUN Treebank & Freq. & \cellcolor[rgb]{1.00,1.00,0.70} \textbf{59.5} & \cellcolor[rgb]{1.00,1.00,0.97} 5.9 & \cellcolor[rgb]{1.00,1.00,0.98} 4.5 & \cellcolor[rgb]{1.00,1.00,0.89} \textbf{22.4} & \cellcolor[rgb]{1.00,1.00,0.97} 6.8 & \cellcolor[rgb]{1.00,1.00,1.00} 0.9 \\
 & Arg. Str. Tran. & $\Delta$ & \cellcolor[rgb]{0.60,0.75,1.00} 1.98 & \cellcolor[rgb]{0.63,0.77,1.00} 1.92 & \cellcolor[rgb]{0.70,0.81,1.00} 1.81 & \cellcolor[rgb]{0.76,0.85,1.00} 1.70 & \cellcolor[rgb]{0.77,0.86,1.00} 1.68 & \cellcolor[rgb]{0.73,0.83,1.00} 1.75 \\
 & Arg. Str. Ditr. & $\Delta$ & \cellcolor[rgb]{0.70,0.81,1.00} 1.81 & \cellcolor[rgb]{0.84,0.90,1.00} 1.56 & \cellcolor[rgb]{0.90,0.93,1.00} 1.47 & \cellcolor[rgb]{0.81,0.88,1.00} 1.62 & \cellcolor[rgb]{0.91,0.94,1.00} 1.45 & \cellcolor[rgb]{1.00,1.00,1.00} 1.29 \\
 
\midrule

\multirow{2}{*}{\rotatebox{90}{\tiny EuroLLM}} & Arg. Str. Tran. & $\Delta$ & \cellcolor[rgb]{0.79,0.87,1.00} 5.24 & \cellcolor[rgb]{0.93,0.95,1.00} 3.40 & \cellcolor[rgb]{0.97,0.98,1.00} 2.77 & \cellcolor[rgb]{0.85,0.91,1.00} 4.39 & \cellcolor[rgb]{1.00,1.00,1.00} 2.42 & \cellcolor[rgb]{0.97,0.98,1.00} 2.83 \\
 & Arg. Str. Ditr. & $\Delta$ & \cellcolor[rgb]{0.60,0.75,1.00} 7.83 & \cellcolor[rgb]{0.73,0.83,1.00} 6.13 & \cellcolor[rgb]{0.76,0.85,1.00} 5.66 & \cellcolor[rgb]{0.65,0.78,1.00} 7.11 & \cellcolor[rgb]{0.79,0.87,1.00} 5.30 & \cellcolor[rgb]{0.88,0.93,1.00} 4.04 \\
 
\midrule
\multirow{2}{*}{\rotatebox{90}{\tiny BERTurk}} & Arg. Str. Tran. & $\Delta$ & \cellcolor[rgb]{0.67,0.80,1.00} 13.41 & \cellcolor[rgb]{0.79,0.87,1.00} 10.24 & \cellcolor[rgb]{0.73,0.83,1.00} 11.92 & \cellcolor[rgb]{0.67,0.79,1.00} 13.44 & \cellcolor[rgb]{0.88,0.93,1.00} 7.87 & \cellcolor[rgb]{0.80,0.87,1.00} 10.16  \\
 & Arg. Str. Ditr. & $\Delta$ & \cellcolor[rgb]{0.60,0.75,1.00} 15.33 & \cellcolor[rgb]{0.84,0.90,1.00} 8.94 & \cellcolor[rgb]{0.76,0.85,1.00} 11.21 & \cellcolor[rgb]{0.64,0.78,1.00} 14.18 & \cellcolor[rgb]{0.88,0.93,1.00} 7.86 & \cellcolor[rgb]{1.00,1.00,1.00} 4.74 \\
\bottomrule
\end{tabular}}
\caption{Word order performance comparison between human judgments and best models. The white-blue gradient represents mean acceptability differences (low to high) for each row, while the white-yellow gradient reflects corpus frequency. Frequency values represent the percentage of different word orders within each corpus. For human judgments, mean differences are calculated using z-score–transformed acceptability ratings, whereas for model evaluation, the mean difference reflects the difference in sequence log probabilities.}
\label{tab:word_order}
\end{table}

Although SOV is the canonical word order in Turkish, \citet{SLOBIN1982229} found that 52\% of utterances in their spontaneous adult speech corpus deviate from this order. Similarly, \citet{10.1007/s10579-021-09558-0} reported that only 59.5\% of sentences in the BOUN Universal Dependencies Treebank follow SOV. Notably, they identified two different word orders as the second most frequent, highlighting how Turkish word order patterns can vary largely between spoken and written language. Both studies, however, agree that VOS is the least attested. 

Native-speaker acceptability judgments reflect that SOV had the highest mean rating difference for both transitive and ditransitive sentences, in line with spoken and written corpus frequencies. The second-highest mean acceptability rating difference for transitive paradigms was the SVO word order, while it was OVS for the ditransitive ones. These are also the second-most-frequent word orders reported by \citet{SLOBIN1982229} and \citet{10.1007/s10579-021-09558-0} respectively. In transitive sentence ratings, VOS is not found to be the most challenging word order, contrary to what might be expected based on attested corpus statistics. This suggests that a rare word order does not inherently hinder people's ability to identify acceptable sentences. Speakers seem to tolerate non-canonical word orders more readily in transitives than in ditransitives. One interpretation may be that case differences are easier to spot in transitive sentences due to fewer arguments.

We see the opposite trend for model evaluations with EuroLLM being particularly sensitive to non-canonical word orders in transitive sentences. BERTurk remains robust to all word orders, showing only a pronounced drop for the rare VOS paradigm in the ditransitive condition. For both transitive and ditransitive sentences, models show high mean log probability differences on OVS word orders. This suggests that model performances align more closely with word order statistics from the BOUN treebank than with those from the spoken language corpus by \citet{SLOBIN1982229}.

\subsection{Effect of Subordination}

Table \ref{tab:combined_subordination_heatmap} displays human and model performance on four subordination paradigms compared to a non-subordinated baseline. In Turkish, subordinate clauses can be finite or non-finite. However, finite subordinate clauses are much less frequent than non-finite ones \citep{göksel2005turkish}. For non-finite subordination, we consider three different subordinating suffixes: -DIK, -(y)IncA, and -(y)ken. -DIK forms nominal subordinate clauses while the latter two form adverbial ones.

The acceptability judgment task appears to be easier in non-finite -DIK subordinates than in finite ones, consistent with finite clauses' lower frequency. While -DIK's mean difference nearly matches the baseline in transitives, it shows a decline for ditransitives.
Among non-finite structures, -(y)IncA and \mbox{-(y)ken} prove harder than -DIK, suggesting that adverbial clauses pose greater challenges. However, performance deficits may also reflect semantic incongruities from augmentation. Some verb roots may conflict with the aspectual property of the adverbial markers. Therefore, we cannot reliably claim inherent difficulty in adverbial clauses.

\begin{table}[ht]
\centering
\small
\adjustbox{max width=0.48\textwidth}{%
\renewcommand{\arraystretch}{1.2}
\begin{tabular}{clrrrrr}
\toprule
 &  & Baseline & Finite & -DIK & -(y)IncA & -(y)ken  \\
\midrule

\multirow{2}{*}{\rotatebox{90}{\scriptsize Human}} 
& Tran. $\Delta$  & \cellcolor[rgb]{0.61,0.76,1.00} 1.97 & \cellcolor[rgb]{0.75,0.84,1.00} 1.60 & \cellcolor[rgb]{0.62,0.76,1.00} 1.95 & \cellcolor[rgb]{0.98,0.99,1.00} 0.98 & \cellcolor[rgb]{1.00,1.00,1.00} 0.92 \\
& Ditr. $\Delta$ & \cellcolor[rgb]{0.60,0.75,1.00} 2.00 & \cellcolor[rgb]{0.87,0.92,1.00} 1.27 & \cellcolor[rgb]{0.75,0.84,1.00} 1.59 & \cellcolor[rgb]{0.83,0.89,1.00} 1.38 & \cellcolor[rgb]{0.94,0.96,1.00} 1.08 \\

\midrule

\multirow{2}{*}{\rotatebox{90}{\tiny EuroLLM}} 
& Tran. $\Delta$ & \cellcolor[rgb]{0.73,0.83,1.00} 5.34 & \cellcolor[rgb]{0.98,0.99,1.00} 2.64 & \cellcolor[rgb]{0.84,0.90,1.00} 4.20 & \cellcolor[rgb]{0.95,0.97,1.00} 2.97 & \cellcolor[rgb]{1.00,1.00,1.00} 2.39 \\
& Ditr. $\Delta$ & \cellcolor[rgb]{0.60,0.75,1.00} 6.83 & \cellcolor[rgb]{0.78,0.86,1.00} 4.83 & \cellcolor[rgb]{0.69,0.81,1.00} 5.85 & \cellcolor[rgb]{0.74,0.84,1.00} 5.31 & \cellcolor[rgb]{0.79,0.87,1.00} 4.73 \\

\midrule

\multirow{2}{*}{\rotatebox{90}{\tiny BERTurk}} 
& Tran. $\Delta$ & \cellcolor[rgb]{0.69,0.80,1.00} 12.90 & \cellcolor[rgb]{1.00,1.00,1.00} 8.47 & \cellcolor[rgb]{0.74,0.84,1.00} 12.10 & \cellcolor[rgb]{0.91,0.95,1.00} 9.70 & \cellcolor[rgb]{0.83,0.90,1.00} 10.81 \\
& Ditr. $\Delta$ & \cellcolor[rgb]{0.60,0.75,1.00} 14.14 & \cellcolor[rgb]{0.93,0.96,1.00} 9.45 & \cellcolor[rgb]{0.65,0.78,1.00} 13.43 & \cellcolor[rgb]{0.73,0.83,1.00} 12.35 & \cellcolor[rgb]{0.77,0.86,1.00} 11.74 \\

\bottomrule
\end{tabular}}
\caption{Subordination performance comparison between human judgments and best models.}
\label{tab:combined_subordination_heatmap}
\end{table}

With human judgment patterns established, we evaluate model performance. EuroLLM’s -DIK performance shows a drop from baseline in transitive sentences. BERTurk mirrors human trends more closely, exhibiting a greater decline in ditransitives. Both models struggle more with finite subordination than -DIK, though EuroLLM shows a sharper contrast. Compared to nominal subordination, both models show smaller mean differences with adverbial clauses. 
Overall, we observe that models show sensitivity to different subordination structures.

\section{Conclusion}

TurBLiMP\footnote{\texttt{\footnotesize https://github.com/ezgibasar/turblimp}} provides the first comprehensive evaluation of language models' syntactic capabilities for Turkish. We find that larger model sizes generally correlate with higher accuracy, with some exceptions. Considerably smaller monolingual language models often outperform their larger multilingual counterparts and perform on par with the best multilingual models. This finding corroborates patterns attested in other syntactic benchmarks \cite{taktasheva-etal-2024-rublimp, jumelet2025multiblimp10massivelymultilingual}. The strong performance of monolingual models highlights the importance of language-specific training for reliable models. Cases where smaller models outperformed larger ones also suggest that scaling alone cannot explain model behavior as far as linguistic evaluations are concerned.

The persistent challenges in phenomena like Ellipsis show that models of all sizes and architectures can struggle with some linguistic phenomena. The discrepancy between model behavior and human judgments for this phenomenon indicates that even the best-performing LLMs may fail to fully capture human linguistic intuition.

TurBLiMP also introduces experimental paradigms to test model robustness to specific linguistic parameters, namely word order and subordination. 
Results on these paradigms reveal subtle sensitivities in high-performing models that standard evaluations would miss, indicating that even models excelling on general minimal pair tasks can exhibit brittleness with the introduction of non-canonical word orders or subordination. 
While some performance patterns align with human judgments, we observe both variation across models and cases of divergence from human judgments.
Furthermore, the human judgments themselves offer valuable insights into native speaker patterns across different subordination structures and word orders, making TurBLiMP a valuable starting point for future research.

In sum, TurBLiMP provides a valuable resource to assess various linguistic phenomena in a controlled fashion, many of which are not represented in prior syntactic benchmarks. 
We hope our work will facilitate linguistically informed model developments and contribute to a better understanding of how language models handle linguistic structures across typologically different languages.

\section*{Limitations}

While TurBLiMP offers a comprehensive evaluation of key linguistic phenomena in Turkish, there are also several limitations to acknowledge. 
In this paper, we evaluated minimal pair acceptability using sequence log probabilities for each sentence. However, our approach represents only one of several valid methods for assessing language models on acceptability benchmarks. For example, \citet{song2025languagemodelsfailintrospect} show that prompting for `meta-linguistic' grammaticality judgments can result in better performance than comparing string probabilities directly.
However, they do show that this `introspective' approach has its limitations, and the optimal way of evaluating linguistic ability in LMs remains an open debate \citep{doi:10.1073/pnas.2400917121}.

\citet{warstadt-etal-2020-blimp-benchmark} define paradigms as minimal pair types and phenomena as broader linguistic categories. Unlike other BLiMP benchmarks that include multiple paradigms under each phenomenon, TurBLiMP currently includes only one paradigm per phenomenon (with the exception of Argument Structure). Splitting some phenomena into multiple paradigms could enable more granular assessment. For a broader coverage of linguistic structures in Turkish, future work could also incorporate additional phenomena into the benchmark.

Our evaluation did not systematically test all available model sizes for each model, which may limit the generalizability of our findings as far as model size is concerned. Testing a wider range of model sizes would strengthen our insights.

Some of the models we tested had both base and instruction-tuned variants. We opted to exclude instruction-tuned versions based on our assumption that English-based tuning was unlikely to improve performance on Turkish grammatical evaluation tasks. Prior work \cite{chirkova-nikoulina-2024-zero-shot} has shown that instruction tuning in English can degrade fluency in non-English languages, and we also observed performance losses when experimenting with the instruction-tuned counterparts of several models. For instance, the average accuracy of Gemma 3 (4B) Instruct was 4.4\% lower than its base variant. That said, this experimental choice limits our ability to comment on potential transfer effects from instruction tuning.

Our word order experiments focused exclusively on sentences with explicit subjects, omitting pro-drop constructions. Given \citet{10.1007/s10579-021-09558-0}'s finding that subjectless sentences exceed the canonical SOV order frequency in the BOUN treebank, future work should investigate these prevalent but untested configurations.

\section*{Acknowledgments}

We are thankful to the anonymous reviewers for their helpful comments and to Ümit Atlamaz for his assistance in recruiting participants. We also thank the survey participants for their time and insight. Ezgi Ba\c{s}ar received funding from the Erasmus Mundus Masters Programme in Language and Communication Technologies, EU grant no. 2019-1508. Francesca Padovani, Jaap Jumelet and Arianna Bisazza were supported by the Talent Programme of the Dutch Research Council (grant VI.Vidi.221C.009).


\bibliography{turblimp}

\appendix

\section{Experimental Paradigm Details}
\label{sec:appendix}

Our experimental paradigms only target minimal pairs belonging to Argument Structure Transitive and Argument Structure Ditransitive phenomena. Below is an example of the transitive baseline.

\begin{exe}
\ex
\begin{xlist}
        \ex
        {
        \gll Bu {[}\underline{şarkı-ya} /*\underline{şarkı-yı}{]} bayıl-ıyor-um. \\
        this {[}song\textsc{\tiny-DAT} /*song\textsc{\tiny-ACC}{]} love\textsc{\tiny-PROG-1SG}\\
        \glt `I love this song.'
        }
\end{xlist}
\end{exe}

For \textbf{word order}, we took the 100 manually crafted minimal pairs from the Transitive and Ditransitive Argument Structure phenomena and, for each phenomenon, we generated 600 pairs for all six possible subject-verb-object permutations. While the original sentences predominantly followed the default Turkish SOV order, the subjects were often omitted and left implicit. We explicitly reintroduced subjects in a dedicated SOV variant and derived the remaining five orders from this augmented set. 

\begin{exe}
\ex
\begin{xlist}
\small
        \ex Ben bu {[}\underline{şarkıya} /*\underline{şarkıyı}{]} bayılıyorum. (SOV)
        \ex Ben bayılıyorum bu {[}\underline{şarkıya} /*\underline{şarkıyı}{]}. (SVO)
        \ex Bu {[}\underline{şarkıya} /*\underline{şarkıyı}{]} ben bayılıyorum. (OSV)
        \ex Bu {[}\underline{şarkıya} /*\underline{şarkıyı}{]} bayılıyorum ben. (OVS)
        \ex Bayılıyorum ben bu {[}\underline{şarkıya} /*\underline{şarkıyı}{]}. (VSO)
        \ex Bayılıyorum bu {[}\underline{şarkıya} /*\underline{şarkıyı}{]} ben. (VOS)
\end{xlist}
\end{exe}

\begin{table}[ht]
\centering
\small
\adjustbox{max width=0.48\textwidth}{%
\renewcommand{\arraystretch}{1.2}
\setlength{\tabcolsep}{3pt}
\begin{tabular}{lllrrrrrr}
\toprule
\multicolumn{1}{c}{} & Data & Metric & SOV & SVO & OSV & OVS & VSO & VOS \\
\midrule

\multirow{2}{*}{\rotatebox{90}{\tiny EuroLLM}} & Arg. Str. Tran. & Acc. & \cellcolor[rgb]{0.43,0.72,0.42} 97.0 & \cellcolor[rgb]{0.50,0.76,0.46} 91.0 & \cellcolor[rgb]{0.62,0.82,0.54} 81.0 & \cellcolor[rgb]{0.53,0.77,0.48} 89.0 & \cellcolor[rgb]{0.61,0.81,0.53} 82.0 & \cellcolor[rgb]{0.58,0.80,0.52} 84.0 \\
 & Arg. Str. Ditr. & Acc. & \cellcolor[rgb]{0.42,0.72,0.41} 98.0 & \cellcolor[rgb]{0.43,0.72,0.42} 97.0 & \cellcolor[rgb]{0.49,0.75,0.45} 92.0 & \cellcolor[rgb]{0.46,0.74,0.44} 94.0 & \cellcolor[rgb]{0.45,0.73,0.43} 95.0 & \cellcolor[rgb]{0.56,0.79,0.50} 86.0 \\

\midrule

\multirow{2}{*}{\rotatebox{90}{\tiny BERTurk}} & Arg. Str. Tran. & Acc. & \cellcolor[rgb]{0.40,0.71,0.40} 99.0 & \cellcolor[rgb]{0.42,0.72,0.41} 98.0 & \cellcolor[rgb]{0.40,0.71,0.40} 99.0 & \cellcolor[rgb]{0.39,0.71,0.39} 100.0 & \cellcolor[rgb]{0.44,0.73,0.42} 96.0 & \cellcolor[rgb]{0.44,0.73,0.42} 96.0 \\
 & Arg. Str. Ditr. & Acc. & \cellcolor[rgb]{0.42,0.72,0.41} 98.0 & \cellcolor[rgb]{0.44,0.73,0.42} 96.0 & \cellcolor[rgb]{0.44,0.73,0.42} 96.0 & \cellcolor[rgb]{0.42,0.72,0.41} 98.0 & \cellcolor[rgb]{0.45,0.73,0.43} 95.0 & \cellcolor[rgb]{0.71,0.86,0.60} 74.0 \\
\bottomrule
\end{tabular}}
\caption{Accuracy scores for word order paradigms.}
\label{tab:acc_word_order}
\end{table}

For \textbf{subordination}, we augmented each paradigm by creating subordinate clauses with three different subordinating suffixes (-(y)IncA, -(y)ken, -DIK) and included a finite subordination paradigm which preserves the original verb inflection. -DIK forms nominal subordinate clauses, while -(y)IncA and -(y)ken form adverbial subordinate clauses. -DIK nominalization carries agreement and case suffixes, and -(y)ken attaches to verb stems marked for aspect.

\begin{exe}
\ex
\begin{xlist}
\small
        \ex Finite \\[2pt]
            Bu {[}\underline{şarkıya} /*\underline{şarkıyı}{]} \textbf{bayılıyorum} sanıyor. \\[2pt]
            `(S)he thinks that I love this song.'
        \ex -DIK \\[2pt]
        Bu {[}\underline{şarkıya} /*\underline{şarkıyı}{]} \textbf{bayıldığımı} sanıyor. \\[2pt]
        `(S)he thinks that I love this song.'
        \ex -(y)IncA \\[2pt]
        Bu {[}\underline{şarkıya} /*\underline{şarkıyı}{]} \textbf{bayılınca} gitti. \\[2pt]
        `$\sim$(S)he left when I really liked this song.' 
        \ex -(y)ken\\[2pt]
        Bu {[}\underline{şarkıya} /*\underline{şarkıyı}{]} \textbf{bayılırken} gitti. \\[2pt]
        `$\sim$(S)he left while I was loving this song.' 
\end{xlist}
\end{exe}

These strategies show varying degrees of morphological complexity, with finite subordination using 2 morphemes, -DIK using 3, -(y)IncA using 1, and -(y)ken using 2. The augmentation procedure yields 400 transitive and 400 ditransitive subordination minimal pairs.

\begin{table}[ht]
\centering
\small
\adjustbox{max width=0.48\textwidth}{%
\renewcommand{\arraystretch}{1.2}
\begin{tabular}{clrrrrr}
\toprule
 &  & Baseline & Finite & -DIK & -(y)IncA & -(y)ken  \\
\midrule

\multirow{2}{*}{\rotatebox{90}{\tiny EuroLLM}} 
& Tran. Acc.  &  \cellcolor[rgb]{0.42,0.72,0.41} 97.6 & \cellcolor[rgb]{0.61,0.81,0.53} 82.0 & \cellcolor[rgb]{0.53,0.77,0.48} 89.0 & \cellcolor[rgb]{0.53,0.77,0.48} 89.0 & \cellcolor[rgb]{0.67,0.84,0.57} 77.0 \\
& Ditr. Acc. & \cellcolor[rgb]{0.43,0.72,0.42} 96.7 & \cellcolor[rgb]{0.53,0.77,0.48} 89.0 & \cellcolor[rgb]{0.43,0.72,0.42} 97.0 & \cellcolor[rgb]{0.46,0.74,0.44} 94.0 & \cellcolor[rgb]{0.54,0.78,0.49} 88.0 \\

\midrule

\multirow{2}{*}{\rotatebox{90}{\tiny BERTurk}} 
& Tran. Acc.  & \cellcolor[rgb]{0.40,0.71,0.40} 99.1 & \cellcolor[rgb]{0.43,0.72,0.42} 97.0 & \cellcolor[rgb]{0.42,0.72,0.41} 98.0 & \cellcolor[rgb]{0.42,0.72,0.41} 98.0 & \cellcolor[rgb]{0.43,0.72,0.42} 97.0 \\
& Ditr. Acc. & \cellcolor[rgb]{0.44,0.73,0.42} 96.1 & \cellcolor[rgb]{0.50,0.76,0.46} 91.0 & \cellcolor[rgb]{0.45,0.73,0.43} 95.0 & \cellcolor[rgb]{0.44,0.73,0.42} 96.0 & \cellcolor[rgb]{0.44,0.73,0.42} 96.0 \\

\bottomrule
\end{tabular}}
\caption{Accuracy scores for subordination paradigms.}
\label{tab:acc_subordination_heatmap}
\end{table}

\section{Acceptability Judgment Collection}
\label{sec:appendix_survey}
\begin{figure}[ht]
  \includegraphics[width=0.48\textwidth]{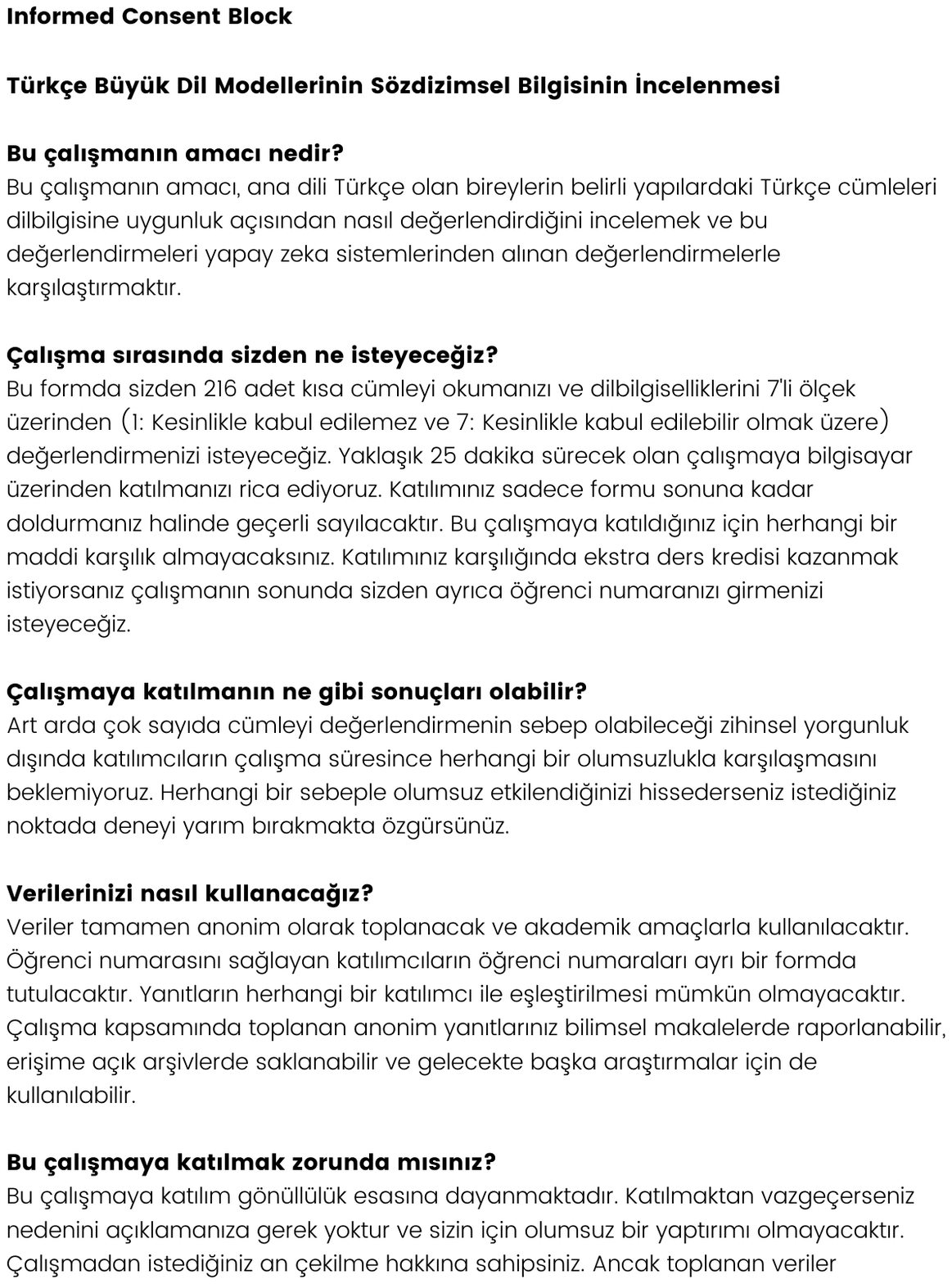}
  \includegraphics[width=0.48\textwidth]{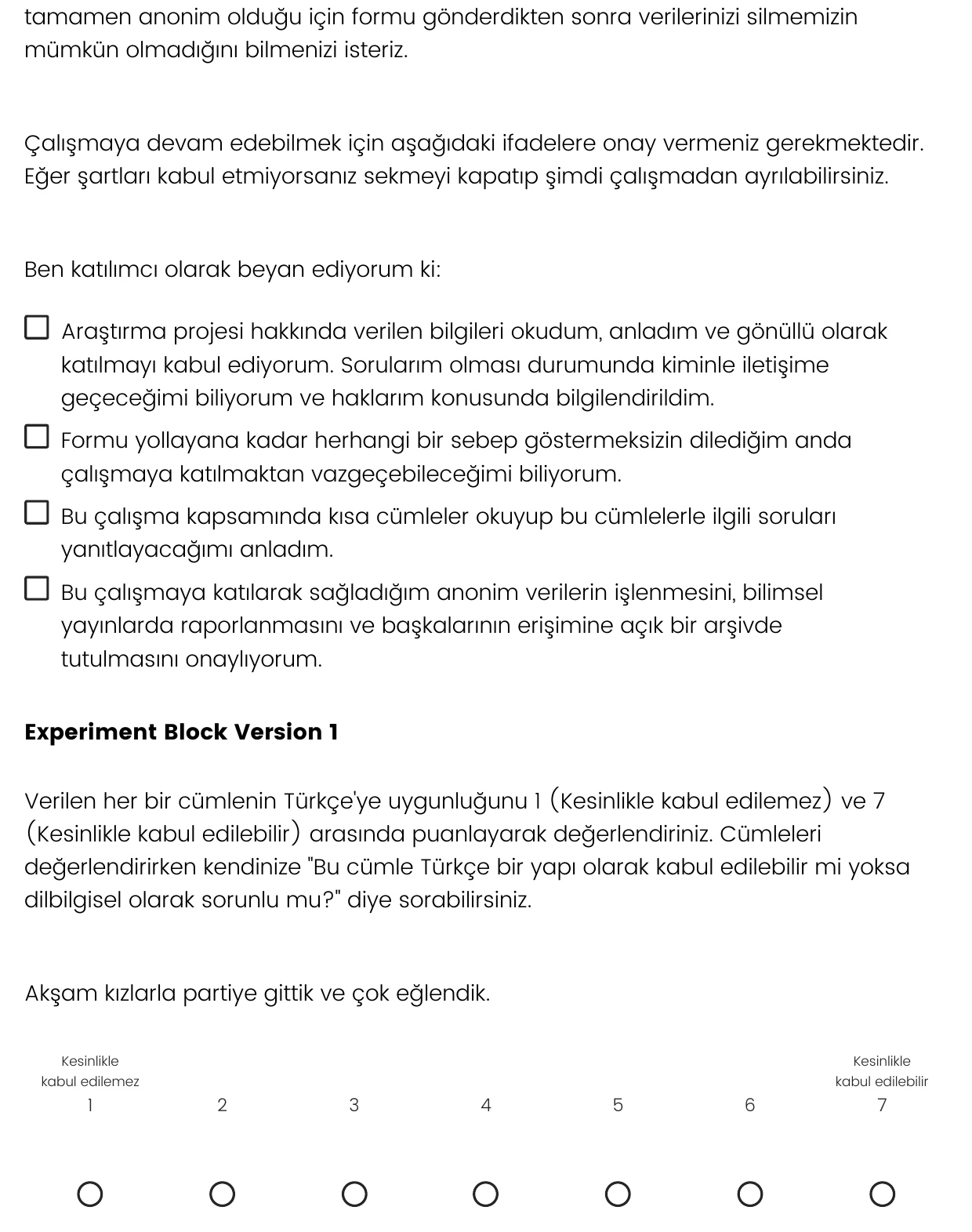}
  \caption{Informed consent form and instructions.}
  \label{fig:survey}
\end{figure}

Figure \ref{fig:survey} provides a screenshot of the survey carried out on the Qualtrics platform. All participants gave their informed consent before starting the survey and agreed that their anonymous responses can be made publicly available. Students received extra credit in exchange for their participation. Participation was on a voluntary basis for both students and non-students, and they were explicitly informed that no financial compensation would be provided. The survey was distributed within our close networks.

\clearpage

\section{Acceptability Judgment Statistics}
\label{sec:appendix_prop}

\begin{table}[ht]
\centering
\begin{tabular}{lr}
\toprule
Category & Proportion \\
\midrule
Transitive SOV & 30/30 \\
Transitive SVO & 30/30 \\
Transitive OSV & 30/30 \\
Transitive OVS & 29/30 \\
Transitive VSO & 30/30 \\
Transitive VOS & 30/30 \\
\midrule
Ditransitive SOV & 30/30 \\
Ditransitive SVO & 30/30 \\
Ditransitive OSV & 28/30 \\
Ditransitive OVS & 28/30 \\
Ditransitive VSO & 30/30 \\
Ditransitive VOS & 29/30 \\
\bottomrule
\end{tabular}
\caption{Proportion of participants with higher average acceptability ratings for acceptable versus unacceptable sentences per word order paradigm.}
\label{tab:prop_word_order}
\end{table}

\begin{table}[ht]
\centering
\begin{tabular}{lr}
\toprule
Category & Proportion \\
\midrule
Transitive Baseline & 30/30 \\
Transitive Finite & 30/30 \\
Transitive -DIK & 30/30 \\
Transitive -(y)IncA & 24/30 \\
Transitive -(y)ken & 28/30 \\
\midrule
Ditransitive Baseline & 29/30 \\
Ditransitive Finite & 29/30 \\
Ditransitive -DIK & 30/30 \\
Ditransitive -(y)IncA & 29/30 \\
Ditransitive -(y)ken & 28/30 \\
\bottomrule
\end{tabular}
\caption{Proportion of participants with higher average acceptability ratings for acceptable versus unacceptable sentences per subordination paradigm.}
\label{tab:prop_subordination}
\end{table}

\begin{table}[ht]
\centering
\begin{tabular}{lr}
\toprule
Category & Proportion \\
\midrule
Anaphor Agreement & 30/30 \\
Argument Structure Tran. & 30/30 \\
Argument Structure Ditr. & 29/30 \\
Binding & 30/30 \\
Determiners & 28/30 \\
Ellipsis & 27/30 \\
Irregular Forms & 30/30 \\
Island Effects & 26/30 \\
Nominalization & 30/30 \\
NPI Licensing & 30/30 \\
Passives & 29/30 \\
Quantifiers & 30/30 \\
Relative Clauses & 30/30 \\
Scrambling & 30/30 \\
Subject Agreement & 30/30 \\
Suspended Affixation & 30/30 \\
\bottomrule
\end{tabular}
\caption{Proportion of participants with higher average acceptability ratings for acceptable versus unacceptable sentences per phenomenon.}
\label{prop:base}
\end{table}

\clearpage

\section{SLOR-based Human Correlations}
\label{sec:slor}

\begin{table}[ht]
\centering
\begin{tabular}{lrr}
\toprule
\textbf{Model} & \textbf{Pearson's $r$} & \textbf{$p$-value} \\
\midrule
Goldfish 1000MB & 0.384 & 0.1422 \\
BERTurk & \textbf{0.557} & \textbf{0.0251} \\
CosmosGPT & 0.414 & 0.1105 \\
Gemma 3 (4B) & 0.400 & 0.1247 \\
Qwen 2.5 & 0.233 & 0.3845 \\
Llama 3.1 & 0.283 & 0.2881 \\
Aya Expanse & 0.454 & 0.0771 \\
Gemma 2 & 0.358 & 0.1730 \\
EuroLLM & 0.357 & 0.1752 \\
Gemma 3 (12B) & 0.385 & 0.1411 \\
\bottomrule
\end{tabular}
\caption{SLOR-based correlation coefficients for various models and their corresponding $p$-values. The statistically significant result ($p < 0.05$) is indicated in boldface.}
\label{tab:slorcorr}
\end{table}

\section{Phenomenon-Agnostic Factors}

\label{sec:phenomenon_agnostic}

\begin{figure*}[ht]
  \centering
  \includegraphics[width=0.8\textwidth]{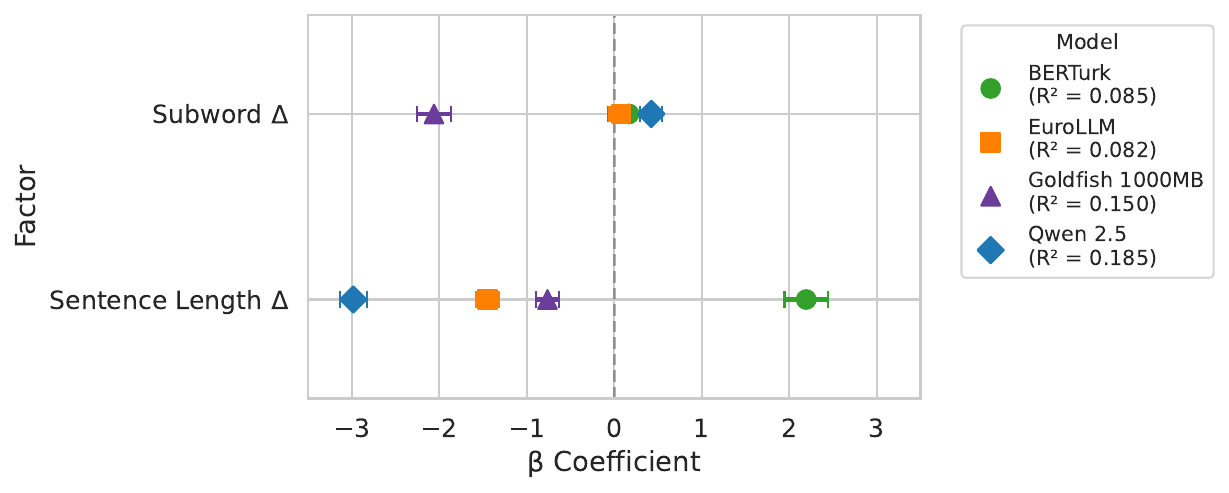}
  \caption{$\beta$ coefficients fitted for the BERTurk, EuroLLM, Goldfish, and Qwen 2.5 models with sentence length and subword count differences as the predictors.}
  \label{fig:regression}
\end{figure*}

To understand general factors influencing model performance beyond phenomena-specific evaluations, we analyzed two potential predictors across all minimal pairs: (1) the difference in sentence lengths between acceptable and unacceptable sentences, and (2) the difference in subword counts for acceptable and unacceptable cue words.

Subword counts refer to the number of smaller units, or subwords, into which a word is segmented by a tokenizer. Depending on the tokenizer design, the resulting subwords may approximate word or morpheme boundaries, but they typically do not perfectly align with either. The subword count difference metric captures the disparity in tokenization granularity between the acceptable and unacceptable cue words and shows whether the acceptable word is split into fewer or more subword units than the unacceptable word.

Using these two predictors, we conduct a linear modeling experiment to predict log probability differences obtained by four different models. We include the two best-performing models (EuroLLM and BERTurk) as well as the two worst-performing models (Goldfish and Qwen 2.5). To allow for better comparability, we opt for the largest Turkish Goldfish model trained on 1000MB of data.

Figure \ref{fig:regression} displays the standardized $\beta$ coefficients for both sentence length differences and subword count differences across all four models. We also include each model's corresponding $R^2$ value, indicating how well these predictors explain the variance in log probability differences.

$R^2$ values are remarkably low for all the models, which means that these two factors alone are not adequate predictors of model behavior. For the purposes of our benchmark, this is the desired outcome as we want model log probabilities to reflect grammatical judgments rather than being confounded by these surface-level features.

\paragraph{Sentence length} Autoregressive and masked models show sentence length effects in opposite directions. In the case of autoregressive models, the model is more likely to incorrectly prefer the unacceptable sentence if the acceptable sentence is longer than its unacceptable counterpart. This indicates that autoregressive models are biased to prefer shorter sequences. BERTurk, which is the only masked language model in our model pool, shows the opposite pattern. As the acceptable sentence gets longer than the unacceptable one, BERTurk is more likely to make the correct prediction. This is possibly due to longer sentences providing more contextual clues for masked token prediction.

Reflecting back on the previously reported accuracy results, we can note that acceptable sentences are typically shorter than unacceptable ones in NPI Licensing, Passives and Quantifiers while the opposite is true in Determiners. Autoregressive models perform better on NPI Licensing, Passives and Quantifiers while BERTurk exhibits slightly better performance for Determiners. This behavioral pattern aligns with the results of our linear model.

\paragraph{Subword counts} Based on prior work \citep{goldman-etal-2024-unpacking,jumelet2025multiblimp10massivelymultilingual}, we expect model performance to degrade with increasing subword count differences. The Goldfish model shows a negative subword count coefficient, behaving in line with our hypothesis. A negative coefficient indicates that the model is more likely to make the wrong prediction if the acceptable cue word is split into more subwords than the unacceptable one. However, contrary to our expectations, all other models show a weak but consistent positive effect. To help us make sense of this counterintuitive finding, we conduct a follow-up analysis examining how each model's tokenizer aligns with morphological boundaries in our stimuli.

Utilizing the morphological inflection pipeline by \citet{akin2007zemberek}, we construct a small dataset of 400 morphologically segmented Turkish words. The dataset is organized into four categories, each targeting morphemes that frequently appear as part of the cue words in our benchmark.

\begin{itemize}
\item \textbf{(Verb + Nominalizer + Possessive + Case)}: Contains 100 verb forms with the nominalizer \textit{-DIK} followed by possessive markers and case endings (e.g., \textit{unuttuğunu} $\rightarrow$ [\texttt{unut}, \texttt{tuğ}, \texttt{un}, \texttt{u}]). This category evaluates the segmentation of nominalized verbs.

\item \textbf{(Verb + TAM + Person)}: Includes 100 finite verbs segmented into stems, tense/aspect/modality (TAM) markers, and person markers (e.g., \textit{olacaktılar} $\rightarrow$ [\texttt{ol}, \texttt{acak}, \texttt{tı}, \texttt{lar}]), targeting inflectional morphology.

\item \textbf{(Noun + (Plural) + Possessive + Case)}: Comprises 100 possessed nouns decomposed into stems, optional plural markers (\textit{-lAr}), possessive markers, and case endings (e.g., \textit{elmalarıma} $\rightarrow$ [\texttt{elma}, \texttt{lar}, \texttt{ım}, \texttt{a}]).

\item \textbf{(Noun + (Plural) + Case)}: Features 100 simpler noun forms with optional plural and case markers (e.g., \textit{elmalara} $\rightarrow$ [\texttt{elma}, \texttt{lar}, \texttt{a}]), providing a baseline for bare nominal inflection.

\end{itemize}

\citet{arnett-bergen-2025-language} previously investigated the morphological alignment of tokenizers across typologically diverse languages to investigate whether a lack of alignment could explain performance gaps in language models. They found no evidence to suggest that morphological alignment plays a significant role. However, their operationalization of morphological alignment differs from ours as far as Turkish is concerned. While they assess whether the tokenizer separates the stem from all other suffixes combined, we investigate whether the tokenizer identifies all the different morpheme boundaries.

We evaluate the models based on their ability to approximate gold-standard morpheme boundaries for the 400 words in our dataset. To that end, we use four different metrics to quantify morphological alignment. These four metrics are: (1) the average Damerau-Levenshtein distance \citep{damerau} between the gold morphemes and the subwords obtained by the model, (2) the proportion of undersegmented words which refers to cases where the model produced fewer segments than the gold data, (3) the proportion of oversegmented words, and (4) the proportion of times when the model produced an output identical to the gold data.

\begin{table}[ht]
\scriptsize
\centering
\begin{tabular}{lrrrr}
\toprule
Tokenizer & Avg. & Undersegm. & Oversegm. & Exact \\
          & Distance         & Items      & Items     & Matches  \\
\midrule
BERTurk    & 2.48            & 82.2           & 0.8           & 5.8 \\
EuroLLM    & 2.36            & 24.8           & 27.5          & 2.5 \\
Goldfish   & 1.62            & 83.0           & 3.5           & 9.8 \\
Qwen 2.5   & 5.50            & 7.0            & 54.2          & 0.5 \\
\bottomrule
\end{tabular}
\caption{Morphological alignment results.}
\label{tab:tokenizer_eval}
\end{table}

Table \ref{tab:tokenizer_eval} illustrates each tokenizer's alignment to gold-standard morpheme boundaries. We observe no correlation between subword count coefficients and the latter three metrics (undersegmentation rate, oversegmentation rate, and exact match proportion). Interestingly, the ranking of models by Damerau-Levenshtein distance perfectly mirrors their ranking by subword count coefficients.

The two best-performing models and the two worst-performing models do not cluster together based on either their subword count coefficients or morphological alignment scores. This reinforces the observation that performance differences cannot be attributed to tokenizer behavior alone.

\end{document}